\documentclass[lettersize,journal]{IEEEtran}
\usepackage{amsmath,amsfonts}
\usepackage{algorithmic}
\usepackage{algorithm}
\usepackage{array}
\usepackage[caption=false,font=normalsize,labelfont=sf,textfont=sf]{subfig}
\usepackage{textcomp}
\usepackage{stfloats}
\usepackage{url}
\usepackage{verbatim}
\usepackage{graphicx}
\usepackage{cite}
\usepackage{dsfont}
\usepackage{amstext}
\usepackage{amsmath}
\usepackage{amssymb}
\usepackage{multirow}
\usepackage{color,xcolor,colortbl}
\usepackage{soul}
\usepackage{booktabs}
\usepackage{makecell}

\definecolor{myblue}{RGB}{186,215,233}
\definecolor{myred}{RGB}{250,162,159}

\usepackage{amsthm}
\theoremstyle{plain}

\theoremstyle{definition}

\theoremstyle{remark}

\usepackage[textsize=tiny]{todonotes}

\newcommand{\tabincell}[2]{\begin{tabular}{@{}#1@{}}#2\end{tabular}}  

\definecolor{shadecolor}{rgb}{0.92,0.92,0.92}
\usepackage{framed}

\begin{document}

\title{Adversarial Robustness in Zero-Shot Learning: \\ An Empirical Study on Class and Concept-Level Vulnerabilities}

\author{
Zhiyuan Peng$^\dagger$, \thanks{\\$^\dagger$: Equal contributions.\\ $^\ddagger$: Corresponding author: Ling Shao (ling.shao@ieee.org).\\
Zhiyuan Peng is with the iFLYTEK Co., Ltd., Hefei 230088, China.\\
Zihan Ye and Ling Shao are with the UCAS-Terminus AI Lab, University of Chinese Academy of Sciences, Beijing 101408, China.\\
Shreyank N Gowda is with the School of Computer Science, the University of Nottingham, NG8 1BB Nottingham, UK. \\
Yuping Yan is with the TGAI lab, the Westlake University, Zhejiang 310030, China. \\
Haotian Xu is with the RippleInfo Co., Ltd, Suzhou 215000, China. \\
}
Zihan Ye$^\dagger$,
Shreyank N Gowda,
Yuping Yan, 
Haotian Xu,
Ling Shao$^\ddagger$,~\IEEEmembership{Fellow,~IEEE},
}
	
%\IEEEpubid{0000--0000/00\$00.00~\copyright~2021 IEEE}
% Remember, if you use this you must call \IEEEpubidadjcol in the second
% column for its text to clear the IEEEpubid mark.
	
\maketitle

\begin{abstract}
Zero-shot Learning (ZSL) aims to enable image classifiers to recognize images from unseen classes that were not included during training. Unlike traditional supervised classification, ZSL typically relies on learning a mapping from visual features to predefined, human-understandable class concepts. While ZSL models promise to improve generalization and interpretability, their robustness under systematic input perturbations remain unclear. In this study, we present an empirical analysis about the robustness of existing ZSL methods at both class-level and concept-level.
Specifically, we successfully disrupted their class prediction by the well-known non-target class attack (clsA).
However, in the Generalized Zero-shot Learning (GZSL) setting, we observe that the success of clsA is only at the original best-calibrated point. After the attack, the optimal best-calibration point shifts, and ZSL models maintain relatively strong performance at other calibration points, indicating that clsA results in a \textit{spurious attack success} in the GZSL.
To address this, we propose the Class-Bias Enhanced Attack (CBEA), which completely eliminates GZSL accuracy across all calibrated points by enhancing the gap between seen and unseen class probabilities.
Next, at concept-level attack, we introduce two novel attack modes: Class-Preserving Concept Attack (CPconA) and Non-Class-Preserving Concept Attack (NCPconA). Our extensive experiments evaluate three typical ZSL models across various architectures from the past three years and reveal that ZSL models are vulnerable not only to the traditional class attack but also to concept-based attacks. These attacks allow malicious actors to easily manipulate class predictions by erasing or introducing concepts.
Our findings highlight a significant performance gap between existing approaches, emphasizing the need for improved adversarial robustness in current ZSL models.
\end{abstract}
	
	
\section{Introduction}
\begin{figure*}[htbp]
    \centering
    \includegraphics[width=0.95\linewidth]{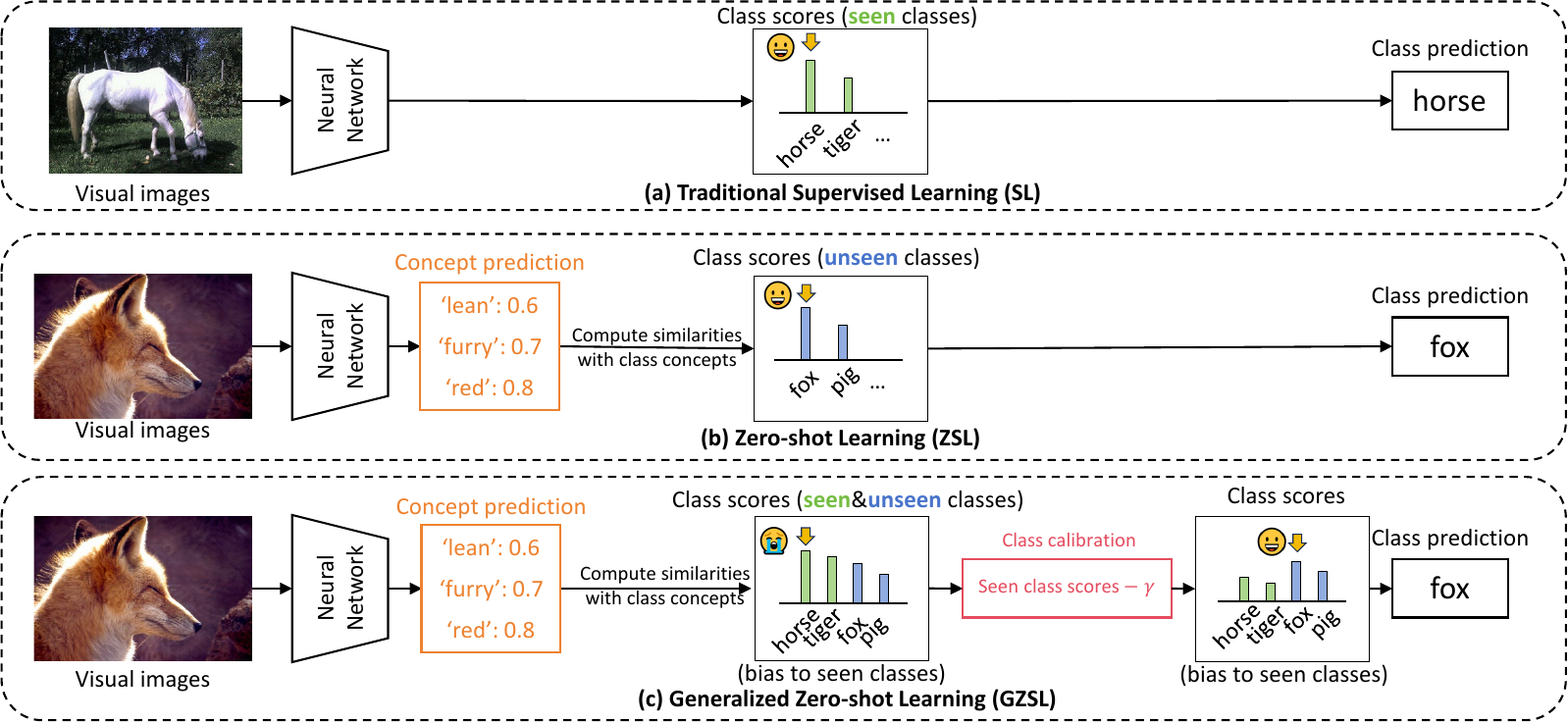}
    \caption{Illustration for the classification process of (a) traditional Supervised Learning (SL), (b) Zero-Shot Learning (ZSL) (c) and Generalized Zero-Shot Learning (GZSL). Different to traditional SL, ZSL methods classify visual images through intermediate concept prediction, i.e. $Visual \to Concept \to Class$. Besides, methods often produce biased class scores in GZSL. Thus, a class calibration technique~\cite{chao2016empirical} is often employed to debias class scores, which uses a pre-defined hyper-parameter $\gamma$ to manually reduce scores of seen classes.}
    \label{fig:banner}
\end{figure*}

Conventional Supervised Learning (SL)~\cite{krizhevsky2012imagenet} has gained significant popularity in a wide range of applications because it can provide an end-to-end solution, directly from image feature extraction to classification. Despite these successes, such methods are highly dependent on the large-scale collection and training of labeled samples, a process that is expensive and time consuming. For example, the popular image classification dataset ImageNet-22k~\cite{ridnik2021imagenetk} contains 14 million images spanning 21,814 classes. Furthermore, conventional supervised methods are particularly sensitive to the number of examples per class; they often fail to accurately classify samples from classes represented by only a few training examples~\cite{wang2020generalizing}.

To reduce the cost associated with collecting and training on labeled samples, the Zero-shot Learning (ZSL) paradigm has been proposed~\cite{lampert2009learning, lampert2013attribute}.
ZSL seeks to enable models to transfer knowledge from seen to unseen classes.
Thus, ZSL training is performed only on images from seen classes, while images from unseen classes are never encountered during training.
To better simulate real-world scenarios, researchers have introduced the generalized ZSL (GZSL) task~\cite{xian2018zero}.
In GZSL, test images originate from both seen and unseen classes, and models must assign labels from a joint label space.
% This setting has attracted considerable attention due to its realistic and challenging nature.

Existing ZSL methods generally leverage various forms of semantic information to facilitate knowledge transfer from seen to unseen classes.
One of the most dominated semantic types is manually defined class concepts (also called attributes)~\cite{lampert2013attribute}.
For examples, the class zebra has concepts `horse-like' and `stripes'.
The common ZSL approach involves predicting the semantic concepts or class concepts based on visual features and then determining the most appropriate class label for each image.
Current research in ZSL often focuses on employing different modules and architectures to extract robust visual features and enhance visual-semantic interactions, e.g. convolutional neural networks (CNNs)~\cite{zhang2017learning, li2018discriminative}, attention modules~\cite{xu2020attribute, ye2023rebalanced}, graph convolutional networks (GCNs)~\cite{xie2020region, guo2023graph}, and Vision Transformers~\cite{naeem2022i2dformer, chen2022transzero, chen2022transzero++}, 
    
% Although these methods have achieved notable progress, they still face challenges such as imbalanced semantic prediction~\cite{} and bias toward seen classes~\cite{}.
%These issues continue to impede the development of a fully trustworthy ZSL framework.

Although these methods have achieved notable progress, they still encounter potential challenges, e.g., the notorious vulnerability to adversarial attacks.
So far, most research on adversarial attacks has focused on supervised learning~\cite{zhang2019adversarial, han2023interpreting}, aiming to improve robustness through various methods.
However, for ZSL, only a limited number of studies have addressed adversarial robustness.
Mere prior work~\cite{yucel2020deep, yucel2022robust, zhang2023atzsl, chen2023zero} investigated the impact of class-specific attacks on ZSL performance.
For example, Harnessing Adversarial Samples (HAS)~\cite{chen2023zero}, enhances ZSL methods by generating adversarial samples via controllable image perturbations that avoid semantic distortion. This scarcity of research on adversarial robustness within the ZSL framework may seriously impede the development of a fully trustworthy ZSL system.

In this paper, we present a comprehensive investigation into the adversarial robustness of ZSL models against adversarial attacks.
As illustrated in Fig.~\ref{fig:banner}, it is important to note that ZSL methods differ significantly from traditional SL approaches in two key aspects:
\begin{enumerate}
    \item \textbf{Concept Prediction}:
    Traditional SL directly derives class predictions from images ($Visual \to Class$).
    In contrast, ZSL models generate class predictions by passing through an intermediate concept prediction stage ($Visual \to Concept \to Class$).
    \item \textbf{Class Bias}: While SL methods have access to labeled data for all classes during training, ZSL methods are trained only on seen classes. Consequently, ZSL models often exhibit a bias toward seen classes, necessitating the use of class calibration techniques~\cite{chao2016empirical} to balance confidences between seen and unseen classes.
\end{enumerate}
Given the additional concept prediction step in ZSL, \emph{both the predicted concepts and the final class predictions must be robust} against adversarial manipulations, especially in sensitive applications such as medical diagnosis~\cite{mahapatra2021medical,rezaei2020zero, mahapatra2022self, gupta2023generative}.

For \textbf{class attacks}, we investigate the security vulnerabilities and robustness of zero-shot learning (ZSL) models against carefully crafted adversarial attacks. First, we conduct an empirical analysis using the well-known class attack (clsA)~\cite{madry2018towards, zhang2025rp} that degrades class predictions by introducing imperceptible noise to input images. However, our further analysis reveals that clsA often results in a \textbf{spurious attack success} in the GZSL setting: while clsA reduces accuracy at the original best-calibrated point, the optimal calibration point subsequently shifts, allowing ZSL models to retain relatively strong performance at alternate points. To resolve this, we propose a Class-Bias Enhancing Attack (CBEA) that, rather than directly disturbing the predicted class probabilities, amplifies the bias between seen and unseen classes. This results in a complete collapse of accuracy across all calibration points.
    
For \textbf{concept attacks}, we conduct an empirical analysis and introduce two novel attack modes: the Class-Preserving Concept Attack (CPconA) and the Non-Class-Preserving Concept Attack (NCPconA). In CPconA, the primary objective is to deteriorate the concept predictions without necessarily altering the final class predictions; in contrast, NCPconA manipulates the intermediate concepts while keeping the final class predictions unchanged. We evaluated three representative methods across various architectures from the past three years, including CNN-based ReZSL~\cite{ye2023rebalanced}, ViT-based PSVMA~\cite{liu2023progressive} and Mamba-based ZeroMamba~\cite{hou2025zeromamba}.
Our extensive experiment demonstrate that ZSL models are vulnerable not only to conventional class attacks but also to these newly proposed concept-based attacks. Such attacks enable adversaries to manipulate class outcomes merely by erasing or introducing specific concepts.

We hope that the framework presented in this study draws the community’s attention to the overlooked issue of adversarial robustness in ZSL models. Moreover, our work serves as a benchmark for future studies and highlights important trends in evaluating ZSL models under adversarial conditions. To better understand real-world performance, we advocate that future research evaluates existing ZSL approaches within adversarial attack settings.

The rest of the paper is organized as follows. Section~\ref{sec:rel} reviews relevant literature. In Section~\ref{sec:attack}, we formally define adversarial attacks on ZSL methods and discuss the inherent challenges. Experimental results, detailed analyses, and discussions are provided in Section~\ref{sec:exp}.
	
\section{Related Work}
\label{sec:rel}

In this section, we first review an overview of ZSL and the untrustworthy drawbacks in ZSL, and then present a review for the progress of adversarial attacks.

\subsection{Zero-shot Learning}
Existing ZSL~\cite{ye2023rebalanced, chen2023explanatory, li2024spiking} can be mainly divided into generative methods and embedding methods. Generative approaches synthesize visual features for unseen classes using deep generative models, e.g. GANs~\cite{xian2018feature,ye2021disentangling, wu2024prototype}, VAEs~\cite{chen2021free, chen2021hsva}, normalizing flows~\cite{shen2020invertible, chen2022gsmflow} and diffusion models~\cite{ye2025zerodiff, gao2025self}.
They use pseudo unseen class examples to transfer ZSL as a traditional supervised learning problem.

We focus on embedding methods~\cite{ye2023rebalanced, liu2023progressive,chen2024progressive}, which typically learn visual-semantic mapping and classify unseen classes in the mapping space.
Classification of unseen classes proceeds by mapping an image into this space and assigning it to the nearest semantic prototype.
In terms of representation of semantics, word2vec/text2vec~\cite{akata2016multi, naeem2022i2dformer, naeem2023i2mvformer} leverages pre-trained language models (e.g. Glove~\cite{pennington2014glove}) to provide continuous semantic labels from online text or class names.
Manually defining concepts~\cite{xu2020attribute, ye2023rebalanced, chen2022transzero, chen2022transzero++, liu2023progressive, du2023boosting, hou2025zeromamba} (also called attributes) is another popular approach.
Vision-based embedding~\cite{xu2022vgse} exploits deep models to mine latent semantics automatically, which are generally considered more discriminative than concepts.
Most recent embedding proposals study ZSL from the model perspective, i.e., by introducing more complex modules (e.g. GCN~\cite{xie2020region, guo2023graph}, attention modules~\cite{xu2020attribute, du2023boosting, chen2024causal}, knowledge distillation~\cite{chen2022transzero++, lu2024pamk}) or frameworks (e.g. Transformer~\cite{liu2023progressive, chen2022transzero,chen2022transzero++}, Mamba~\cite{hou2025zeromamba}) to enhancing the visual–semantic interaction.
However, there are still many key challenges impeding implementing a trustworthy ZSL framework, e.g., class bias and imbalanced semantic prediction.

\textbf{Class Bias.}
Because ZSL models are trained only on seen classes, they exhibit a strong bias toward predicting seen categories at test time, especially under theGZSL setting~\cite{chao2016empirical}. Domain‑aware bias correction methods~\cite{min2020domain} employ regularization or calibrated stacking to mitigate this preference. Generative methods similarly balance seen/unseen decision boundaries by synthesizing features for unseen classes, though bias is seldom fully eradicated.

\textbf{Imbalanced Concept Prediction}
Embedding ZSL often suffers from unequal semantic regression errors: Some concept dimensions are accurately predicted while others are ignored. ReZSL~\cite{ye2019sr} formally analyzed this imbalance and introduced Re‑balanced MSE (ReMSE), which dynamically re‑weights per‑dimension losses based on their error distributions to ensure fair treatment across all semantics.

This work reveals a new untrustworthy drawback: the \textbf{adversarial vulnerability} of ZSL with four comprehensive attack modes for both concepts and classes.
Although a few prior work also discuss about adversarial robustness of ZSL~\cite{yucel2020deep, yucel2022robust, zhang2023atzsl, chen2023zero}, they mainly focus on class-specific attacks.
In contrast, our systematical experiments provide novel insights:
(1) Concept attack: Malicious perturbations can selectively erase or insert mid‑level concepts, leading to misclassification even when final class confidences appear unchanged.
(2) Spurious class attack success: Traditional adversarial class attacks may produce a drop in accuracy only at one calibration point, while ZSL models can recover accuracy by shifting calibration, resulting in a misleadingly high reported robustness.
This work highlights the need for adversarially robust ZSL methods and sets a new benchmark for future research.

\subsection{Adversarial Attack}
Adversarial attacks~\cite{akhtar2018threat} exploit small, often imperceptible, perturbations to inputs in order to induce misclassification. The concept dates back to good word insertion attacks on spam filters, progressed through attacks on linear~\cite{marzi2018sparsity} and kernel‑based classifiers (e.g., SVMs)~\cite{kim2005kernel}, and culminated in gradient‑based methods against deep networks—beginning with Szegedy~\cite{szegedy2013intriguing} and Fast Gradient Sign Method (FGSM)~\cite{goodfellow2014explaining}, followed by multi‑step attacks (BIM)~\cite{kurakin2017adversarial}, Projected Gradient Descent (PGD)~\cite{goodfellow2014explaining}, one‑pixel attacks~\cite{su2019one}, and even perturbations on physical world~\cite{eykholt2018robust}.

Recently, due to the rise of multi-modal learning~\cite{xu2023multimodal, yuan2025survey}, researchers have shifted focus from manipulating classification on single modality to multi-modal models~\cite{cui2024robustness, jiang2025survey, zhou2024revisiting}, especially on attacking intermediate, human-understandable language concepts of classes.
For example, ~\cite{le2024towards} proposed learning more robust feature attributions while \cite{lakkaraju2020robust} try to learn more robust concepts.
And some defense methods \cite{huai2022towards} improve the robustness of the feature-level model explanations.
However, these works concentrate on post‑hoc interpretations of learned concepts rather than the explicit vision–semantic mapping process central to ZSL, marking a fundamental distinction from our research.

\section{Preliminary}
\label{sec:preliminary}

In this section, we begin by rigorously defining the ZSL problem.
We then describe the particular ZSL framework under consideration and detail its training procedure.
Finally, we present the inference mechanisms for both the standard ZSL and the more challenging GZSL settings.
Notably, the class calibration technique (Eq.~\ref{eq:gzsl_inference}) plays a crucial role in controlling bias toward seen classes for a long time.
But in this work, we further demonstrate that it also is pivotal for inducing spurious attack successes in GZSL.

\subsection{Notations}

In the ZSL setting, there are two disjoint label sets: a seen set \( \mathcal{Y}^{s} \) used for training and an unseen set \( \mathcal{Y}^{u} \) used for testing, where \( \mathcal{Y}^{s} \cap \mathcal{Y}^{u} = \emptyset \).
In existing concept-based methods, a set of high-quality concepts $\mathcal{H}$ (e.g., `tails,' `long leg') for all classes is defined by experts~\cite{yang2023language}. Each concept is scored individually for each class, typically based on the number of occurrences~\cite{wah2011caltech}, to obtain the concept-based semantic embedding spaces $\mathcal{S}^{s}_{h}$ for seen classes and $\mathcal{S}^{u}_{h}$ for unseen classes.
Thus, the training dataset is denoted as \( \mathcal{D}^{tr} = \{(\mathbf{x}^{s}, y^{s}, \mathbf{s}^{s}) \mid \mathbf{x}^{s} \in \mathcal{X}^{s}, y^{s} \in \mathcal{Y}^{s}, \mathbf{s}^{s} \in \mathcal{S}^{s} \} \), where \( \mathcal{X}^{s} \), \( \mathcal{S}^{s} \), and \( \mathcal{Y}^{s} \) represent the image, semantic, and label spaces for the seen classes.
The objective in ZSL is to use this training dataset \( \mathcal{D}^{tr} \) to create a classifier capable of predicting unseen classes for images in the test dataset \( \mathcal{D}^{te} = \mathcal{D}^{u} = \{(\mathbf{x}^{u}, y^{u}, \mathbf{s}^{u}_{h}) \mid \mathbf{x}^{u} \in \mathcal{X}^{u}, y^{u} \in \mathcal{Y}^{u}, \mathbf{s}^{u} \in \mathcal{S}^{u} \} \), i.e., \( F_{zsl}: \mathcal{X}^{u} \to \mathcal{Y}^{u} \).
In the GZSL task, test samples may come from both seen and unseen classes. Let $\mathcal{D}^{te,s}$ represent the portion of seen class samples reserved for testing, so the testing dataset becomes \( \mathcal{D}^{te} = \mathcal{D}^{te,s} \cup \mathcal{D}^{u} \). The goal in GZSL is then defined as \( F_{gzsl}: \mathcal{X}^{s} \cup \mathcal{X}^{u} \to \mathcal{Y}^{s} \cup \mathcal{Y}^{u} \).

We focus on attacking Embedding ZSL~\cite{xie2019attentive,liu2021goal} methods.
Generally, embedding ZSL use a feature extracting backbone $f_{bb}$ to extract visual features from a single image $\mathbf{x}$, denoted $\mathbf{v} = f_{bb}(\mathbf{x})$.
The $f_{cls}$ is often a ResNet101~\cite{ye2023rebalanced} or a Visual transformer (ViT)~\cite{liu2023progressive} or Visual Mamba~\cite{hou2025zeromamba}.
Then, a concept prediction module (generally is a fully-connected layer) is employed to predict concept-based class embedding, i.e. $ \tilde{\mathbf{s}} = f_{con}(\mathbf{v})$.
Finally, a class prediction operator $f_{cls}$ (Finds the most similar class in concept space) is conducted by the predicted concept-based class embedding.
In other words, $F_{zsl} = f_{cls} \circ f_{con} \circ f_{bb}$.

\subsection{Target ZSL models} % might could be removed?
In our experiments, various embedding ZSL methods are evaluated.
Due to the compact structure and excellent performance of ReZSL~\cite{ye2023rebalanced}, we take it as a typical example and provide a detailed introduction of it in the part.
Given an input image $\mathbf{x} \in \mathbb{R}^{H\times W\times C}$ where $(H, W)$ is the size of an input image with $C$ as the RGB channels, ReZSL first reshapes it into a sequence of flattened 2D patches.
Let us use $N$ to denote the number of flattened patches.
Then, it uses a pre-trained model, such as ResNet or ViT, as the backbone $f_{bb}$ for extracting  patch-wise visual features $\mathbf{v}_{patch} \in \mathbb{R}^{N \times d_v} $ while $d_v$ is the feature dimension.
Next, ReZSL uses a dual-stream module, a global branch for global visual features and a local branch for semantic-specific features.
Specifically, in the global branch, the patch-wise features $\mathbf{v}_{patch}$ were compressed by global average pooling as $\mathbf{v}_{global} \in \mathbb{R}^{d_v}$.

In the local branch, ReZSL uses Glove (a word2vec model) to obtain the concept embedding $E \in \mathbb{R}^{|\mathcal{H}|\times d_{s}}$ of every concepts $\mathbf{h} \in \mathcal{H}$, while $d_{s}$ is the dimension of concept embeddings.
A cross-modality interaction is utilized to search the visually-relevant concepts by using image patches as the queries.
In other words, let us define $Q = \mathbf{v}_{patch}\times W_q$ as the image queries, $K = E \times W_k$ as the concept keys to compare with, and $V = E\times W_v$ as the concept values to mix with after the search, where $W_q$, $W_k$ and $W_v$ are learnable linear transformations.
The cross-modal attention is estimated by computing the similarity between queries and keys:
\begin{equation}
    A(\mathbf{v}_{patch}, K) = softmax(Q K^\top)
\end{equation}
This attention matrix is used to compute semantic-specific features $\mathbf{v}_{specific} \in \mathbb{R}^{|\mathcal{H}|×d_v}$ for all concepts as linear combinations of rows of the value matrix $V$ i.e.,$ \mathbf{v}_{specific} = A(\mathbf{v}_{patch}, K) \times V$. Intuitively, this operation recomputes the patch features using the relevant concept word2vec.
    
Finally, ReZSL fuses global features and local features by 
\begin{equation}
    \mathbf{v}_{final} = \mathbf{v}_{specific} + \mathbf{\dddot{v}}_{global},
\end{equation}
where $\mathbf{\dddot{v}}_{global} \in \mathbb{R}^{|\mathcal{H}|×d_v}$ is the extended global features to keep the same shape with local features.
Finally, it uses the concept prediction module $f_{con}$ (a fully connected layer) to predict the concept-based class embedding
\begin{equation}
\tilde{\mathbf{s}} = f_{con}(\tilde{\mathbf{v}}_{final}) = \mathbf{W} \times \tilde{\mathbf{v}}_{final},
\end{equation}
where $\mathbf{W}\in \mathbb{R}^{d_{s} \times d_{v}}$ is the parameter matrix of $f_{con}$.

\subsection{Training Loss}
Following prior work~\cite{xu2020attribute,liu2021goal,du2022boosting}, ReZSL employs the widely used Semantic Cross‑Entropy (SCE) loss:
\begin{equation}
    \label{eq:SCE}
    \mathcal{L}_{SCE}(\mathbf{x}, y, \mathcal{Y}_s) = -\log p_{y}(\mathbf{x}|\mathcal{Y}_s),
\end{equation}
where the posterior probability of the true class $y$  for sample $\mathbf{x}$ is defined as
\begin{equation}
    \label{rezsl:eq:CP}
    p_{y}(\mathbf{x}|\mathcal{Y}_s) = \frac{
        e^{ 
            \tau\cos (\tilde{\mathbf{s}}, \mathbf{s}) 
        } 
    } {
        \sum_{y^\prime \in \mathcal{Y}_s }  e^{ 
            \tau\cos (\tilde{\mathbf{s}}, \mathbf{s}_{y^\prime}) 
        }
    } .
\end{equation}
Here, $\tau$ denotes a learnable scaling factor, and $\cos (\tilde{\mathbf{s}}, \mathbf{s}_{y^\prime})$ measures the angular similarity between the predicted semantic vector $\tilde{\mathbf{s}}$ and the semantic prototype $\mathbf{s}_{y^\prime}$ of class $y^\prime$.
Concretely, by normalizing each vector to unit length,
\begin{align}
    \tilde{\mathbf{t}} = \tilde{\mathbf{s}}/\|\tilde{\mathbf{s}}\|_2 ,
    \mathbf{t}_c = \mathbf{s}_c/\|\mathbf{s}_c\|_2,
\end{align}
one obtains $\cos (\tilde{\mathbf{s}}, \mathbf{s}_{y^\prime}) = \tilde{\mathbf{t}}^{T} \mathbf{t}_{y^\prime}, $
which depends solely on the directional alignment of the two vectors. This normalization ensures that the loss focuses on the angular relationship.
Thus, enhancing the model capacity to distinguish between classes based on semantic orientation rather than magnitude.

To mitigate bias in semantic predictions, ReZSL introduces the Rebalanced Mean Squared Error (ReMSE) loss, which enforces a more uniform semantic predictor across classes and dimensions. Let $M^{\prime} \in \mathbb{R}^{|\mathcal{Y}^s| \times d_{s}}$ denote the matrix of class‑averaged semantic errors computed per batch, providing a shared scale for re‑weighting.
For each class label $l$ and semantic dimension $j$, ReMSE defines two complementary weighting factors: (1) a semantic-level re-weight factor $p_{lj}$ that balances different semantic dimensions within the same class, and (2) a class-level re-weight factor $q_{lj}$ that equalizes contributions across different classes for the same semantic.
the semantic-level balancing factor $p_{lj}$ is calculated as follows:
\begin{equation}
    p_{lj} = \left(\log \frac{
        m^{\prime}_{lj}
    }{
        \min_{1 \le k \le d_s} m^{\prime}_{l k}
    } + 1 \right)^{\beta},  \quad  \beta \ge 0.
\end{equation}
A logarithmic function is used here to avoid potential ratio explosions, and a parameter $\alpha$ is taken to control the scale of re-weighting.
Similarly, the class-level balancing factor $q_{lj}$ is designed by
\begin{equation}
    q_{lj} = \left(\log \frac{ 
        m^{\prime}_{lj} 
    }{
        \min_{c \in \mathcal{Y}^s} m^{\prime}_{cj} 
    } + 1 \right)^{\alpha},  \quad  \alpha \ge 0.
\end{equation}
Finally, the ReMSE loss function is:
\begin{equation}
    \label{eq:ReMSE}
    \mathcal{L}_{ReMSE}  = \frac{1}{N} \sum^{N}_{i=1} \sum^{d_{s}}_{j=1} p_{y_i j} q_{y_i j} o_{ij},
\end{equation}
where $ o_{ij}= (\tilde{t}_{ij}-t_{ij})^2$. This re-balancing strategy ensures that both infrequent classes and underrepresented semantic dimensions receive amplified gradients, leading to a more even semantic predictor.

\subsection{Zero‐Shot Inference}
In the ZSL task, each test sample $\mathbf{x} \in \mathcal{X}^u$ is assigned to the unseen class whose prototype best aligns with its projected embedding:
\begin{equation}
\label{eq:zsl_inference}
\tilde{y} = \arg\max_{y^\prime \in \mathcal{Y}^u} \;\cos\!\bigl(\tilde{\mathbf{s}},\,\mathbf{s}_{y^\prime}\bigr).
\end{equation}
Under the more challenging GZSL protocol, where instances may originate from both seen and unseen categories, we expand the search space to $\mathcal{X}^s \cup \mathcal{X}^u$ and correspondingly to $\mathcal{Y}^s \cup \mathcal{Y}^u$.  
To mitigate the well‐known bias toward seen classes, a calibration term is subtracted from the cosine score of every seen category~\cite{chao2016empirical}:
\begin{equation}
\label{eq:gzsl_inference}
\boxed{
\tilde{y} = \arg\max_{y^\prime \in \mathcal{Y}^s \cup \mathcal{Y}^u} \Bigl[\cos\!\bigl(\tilde{\mathbf{s}}, \mathbf{s}_{y^\prime}\bigr) - \textcolor{red}{\gamma} \mathbb{I}[y^\prime\in\mathcal{Y}^s]\Bigr],
}
\end{equation}
where $\mathbb{I}[\cdot]$ is the indicator function and $\gamma$ controls the degree of penalization applied to seen‐class scores.

Intuitively, $\gamma$ interpolates between two extremes. As $\gamma \to +\infty$, all seen‐class scores are driven to $-\infty$, reducing the rule to conventional ZSL over unseen classes only. Conversely, as $\gamma \to -\infty$, the model effectively reverts to a standard closed‐set classifier that only predicts seen classes. By selecting an intermediate $\gamma$, one strikes a balance between the conservative assignment of points to unseen categories and the aggressive pull toward seen categories.
   
\section{Adversarial Attacks against ZSL}
\label{sec:attack}
In this section, we now present four distinct attack modes: (1) traditional Class Attack (clsA), (2) Class-bias Enhancing Attack (CBEA), (3) Non-class-preserving Concept Attack (NCPconA) and (4) Class-preserving Concept Attack (CPconA).
The first two are general enough and can be used to attack any embedding ZSL methods, and the last two modes expose vulnerabilities of any concept‑based model, underscoring the urgent need for robust ZSL.

\subsection{Traditional Class Attack}
To analyze traditional classification robustness, we employ the well-known Projected Gradient Descent (PGD) attack~\cite{goodfellow2014explaining} as our classification attacker (clsA).
Specifically, clsA iteratively updates adversarial examples using the sign of the gradient of the cross-entropy loss.
We denote the attack with \(T\) update steps as clsA-\(T\), and the perturbed image after the \(t\)-th step as
\[
  \mathbf{x}^\star_t = \mathbf{x} + \delta_t,\quad t = 0,\dots,T-1,
\]
where the initial perturbation \(\delta_0\) is drawn uniformly from \(\mathcal{U}(-\epsilon,\epsilon)\).
The update rule at each step in ZSL is:
\begin{align}
\label{eq:clsA_ZSL}
    & \mathbf{x}^{\star}_{t+1} = \mathbf{x}^{\star}_{t} + \alpha \cdot sign(\nabla_{\mathbf{x}} \mathcal{L}_{sce}(\mathbf{x}^{\star}_{t}, y, \mathcal{Y}_u)), 
\end{align}
where \(\mathcal{L}_{\mathrm{sce}} (\mathbf{x}^{\star}_{t}, y, \mathcal{Y}_u)) \) denotes the semantic cross-entropy loss to unseen class set, \(\alpha\) is the step size (often set to \(\epsilon/T\)).
The \(\mathrm{sign}(\cdot)\) function returns \(+1\) for positive elements and \(-1\) for negative ones.
To make the generated adversarial examples satisfy the $L_{\infty}$ bound, one can element-wisely clip $\mathbf{x}^{\star}_{T}$ into the $\epsilon$ vicinity of  $\mathbf{x}$, i.e.
\begin{align}
    \label{eq:constraint}
    ||\mathbf{x}^{\star}_{T}-\mathbf{x}||_{\infty}\leq \epsilon_{thresh},
\end{align}
or simply set $\alpha = \epsilon/T$.
The constraint guarantees that the generated perturbation is imperceptible, so it cannot be easily detected.
It is well established that iterative attacks typically produce stronger white-box adversaries than single-step methods like FGSM~\cite{goodfellow2014explaining}, albeit with reduced black-box transferability~\cite{papernot2016transferability}. 

We evaluate the adversarial robustness by measuring the change in ZSL accuracy (computed via the ZSL inference Eq.~\ref{eq:zsl_inference}) to the final clipped adversarial example \(\mathbf{x}^\star_T\).
For GZSL, we simply use the full class set $\mathcal{Y}_u \cup \mathcal{Y}_s$ to replace the unseen class set of SCE loss in Eq.~\ref{eq:clsA_ZSL}, i.e., 
\begin{align}
\label{eq:clsA_GZSL}
    & \mathbf{x}^{\star}_{t+1} = \mathbf{x}^{\star}_{t} + \alpha \cdot sign(\nabla_{\mathbf{x}} \mathcal{L}_{sce}(\mathbf{x}^{\star}_{t}, y, \mathcal{Y}_u\cup \mathcal{Y}_s)), 
\end{align}
Of course, to evaluate GZSL accuracy, we also need to switch to Eq.~\ref{eq:gzsl_inference} to infer \(\mathbf{x}^\star_T\).

    \subsection{Class-Bias Enhancing Attack}
    In our experiments, we almost completely reduced the GZSL accuracy of the ZSL model using such a simple clsA attack.
    However, with further analysis, we find that, since the calibration technique, i.e. the $ - \gamma \mathbb{I}[y\in\mathcal{Y}^s]$ in Eq.~\ref{eq:gzsl_inference}, ZSL models exhibit a \textbf{spurious adversarial vulnerability}.
    In other word, the GZSL accuracy of ZSL models has almost completely decreased \textbf{only at the current calibration point $\gamma$}.
    If $\gamma$ is shifted elsewhere, ZSL models still maintains a relative GZSL accuracy.

    To this end, we propose the \textbf{Class-Bias Enhancing Attack (CBEA)}.
    Different to clsA (PGD~\cite{goodfellow2014explaining}) and other classification attacks~\cite{goodfellow2014explaining} that directly attack the class probability, our CBEA enhances the class prediction bias between seen and unseen classes by gradient ascent, i.e., 
    \begin{align}
    \label{eq:CEBA}
    & \mathbf{x}^{\star}_{t+1} = \mathbf{x}^{\star}_{t} + \alpha \cdot sign(\nabla_{\mathbf{x}} \mathcal{L}_{CB}(\mathbf{x}^{\star}_{t}, \mathcal{Y}_u,  \mathcal{Y}_s)), 
    \end{align}
    where $\mathcal{L}_{CB}(\mathbf{x}^{\star}_{t}, \mathcal{Y}_u,  \mathcal{Y}_s))$ measures the class prediction bias by computing the difference between averaged cosine similarities of seen classes and unseen classes.
    Its formulation is:
    \begin{align}
    \label{eq:CB}
     \mathcal{L}_{CB}(\mathbf{x}^{\star}_{t}, \mathcal{Y}_u, \mathcal{Y}_s)) & = \frac{1}{|\mathcal{Y}^s|} \sum_{y^\prime \in \mathcal{Y}^s} \cos\!\bigl(\tilde{\mathbf{s}}, \mathbf{s}_{y^\prime}\bigr) \nonumber \\ 
     & - \frac{1}{|\mathcal{Y}^u|} \sum_{y^\prime \in \mathcal{Y}^u} \cos\!\bigl(\tilde{\mathbf{s}}, \mathbf{s}_{y^\prime}\bigr),
    \end{align}
    where $|\mathcal{Y}^s|$ and $|\mathcal{Y}^u|$ are the number of seen classes and unseen classes, respectively.
    
    Although our CBEA does not directly attack class probabilities, it still effectively reduces model accuracy on both seen and unseen classes. More importantly, it will not trigger the spurious adversarial vulnerability.
    No matter how $\gamma$ changes, the GZSL accuracy of models has been reduced to a relatively low level in our experiments.

    \subsection{Non-class-preserving Concept Attack}
    Apart from class prediction, the concept prediction also is a critical component of ZSL models.
    Under certain conditions, malicious attackers might only want to maximize the disruption of the ZSL model's conceptual predictions.
    To evaluate robustness at the concept-level, we propose the Non-class-preserving Concept Attack (NCPconA) to simulate a more comprehensive attack environment, i.e.,
    \begin{align}
    \label{eq:ncpcona}
    & \mathbf{x}^{\star}_{t+1} = \mathbf{x}^{\star}_{t} + \alpha \cdot sign(\nabla_{\mathbf{x}}\mathcal{L}_{MSE}( f_{con}(f_{bb}(\mathbf{x})), f_{con}(f_{bb}(\mathbf{x}^{\star}_{t}))),
    \end{align}
    where $f_{con}(f_{bb}(\mathbf{x}^{\star}_{t})))$ is the generated concept-based explanations to interpret the predicted class label for the crafted adversarial sample $\mathbf{x}^{\star}_{t}$.
    The objective function is to maximize the MSE between generated concepts before and after the attacks.
    In practical settings, NCPconA could be used to remove relevant concepts and introduce non-relevant concepts by perturbing intermediate concept output.

    % Compare to CPconA, the NCPconA only relax the class prediction constraint.
    
    \subsection{Class-preserving Concept Attack}
    The above NCPconA is used to increase the difference between targeted and predicted concept-based embeddings, but it might change class prediction.
    In many cases, a decrease in class predictions may trigger security warnings.
    To this end, we propose another concept attack method, Class-preserving Concept Attack (CPconA), that attacks concept prediction with the constraint of not changing the final class prediction.
    The CPconA is a much more powerful attack than the NCPconA as it also destroys concepts while maintaining the same class prediction.
    This can be especially troublesome as it would defeat the interpretability of ZSL models.
    Its update rule is the following optimization-based adversarial framework:
    \begin{align}
    \label{eq:cpcona}
        \mathbf{x}^{\star}_{t+1} & = \mathbf{x}^{\star}_{t} + \alpha \cdot sign(\nabla_{\mathbf{x}}\mathcal{L}_{MSE}( f_{con}(f_{bb}(\mathbf{x})), f_{con}(f_{bb}(\mathbf{x}^{\star}_{t}))), \nonumber \\
        & \text{subject to} \quad F_{zsl}(\mathbf{x}^{\star}_{T}) = F_{zsl}(\mathbf{x}),
    \end{align}
    where the constraint ensures that the class predictions of the sample $\mathbf{x}$ are identical before and after the attack.

    The above four schemes define the goals of the attacker from four different angles. Based on the above proposed adversarial attacks, we can perform security vulnerability analysis to understand how motivated attackers can craft malicious examples to mislead ZSL models to generate wrong concepts. The magnitude of perturbation reflect the robustness of ZSL models to attacks. The smaller the magnitude of the crafted adversarial perturbations, the less robust the generated concepts are to the adversarial attacks.

\section{Experiments}
To demonstrate the vulnerability of ZSL models, we evaluate various ZSL methods with different backbone under our attack modes on multiple metrics both on ZSL and GZSL settings over three popular benchmark datasets. Our evaluated methods include the ResNet-based ReZSL~\cite{ye2023rebalanced} (TIP23), ViT-based PSVMA~\cite{liu2023progressive} (CVPR23) and Mamba-based ZeroMamba~\cite{hou2025zeromamba} (AAAI25).
We first present the ZSL and GZSL results for the plain class attack clsA in Sec.~\ref{sec:exp_clsA} and empirically exhibit the spurious attack success in GZSL.
Then, we provide results for CBEA that fully remove GZSL accuracies at almost gamma points in Sec.~\ref{sec:exp_CBEA}.
Next, several results for CP-conA and NonCP-conA are provided to show the catastrophic concept disturbance in Sec.~\ref{sec:exp_NCPconA} and ~\ref{sec:exp_CPconA}.
Finally, we provide visualization analysis of our attack modes to show their invisibility in Sec.~\ref{sec:exp_vis}.

\textbf{Datasets}: We conduct extensive experiments to evaluate the proposed attack methods on three different ZSL benchmarks, namely (1) the coarse-grained dataset AWA2~\cite{xian2018zero}, one extensive animal dataset composed of 37,322 images from 50 classes (40 seen and 10 unseen) with 85-dim concepts ranged from 0 to 100 ($-1$ denotes missing data); (2) the fine-grained bird dataset CUB~\cite{wah2011caltech, reed2016learning}, containing 11,788 images in 200 (150 seen and 50 unseen) classes with 312 concepts ranging from 0 to 100; (3) the fine-grained dataset SUN~\cite{patterson2012sun}, a large-scale dataset including 14,340 images from 717 classes (645 seen and 72 unseen) with 102 concepts ranging from 0 to 1. We divide these data into training and testing sets following~\cite{xian2018zero}, which is widely used in present methods.

\textbf{Evaluation Protocols}:
We adopt a variety of metrics for comparison.
Specifically, for class attack modes, we calculate the top-1 classification accuracy ($T1$) of unseen classes in ZSL.
In GZSL, we calculate three kinds of top-1 accuracies, namely the accuracy for unseen classes (denoted as $U$), the accuracy for
seen classes ($S$), and their harmonic mean:
\begin{equation}
    H = \frac{2 \times U \times S}{U+S}.
\end{equation}
Besides, to exhibit effectiveness of our CEBA in all $\gamma$ point, we report the performance based on the Area Under Seen-Unseen accuracy Curve (AUSUC)~\cite{chao2016empirical}, which evaluates the degree of trade-off between $U$ and $S$ in GZSL.
Finally, for concept attack modes,we exploit two new metrics, the MSE of the predicted concepts for unseen ($MSE_u$) and seen ($MSE_s$) classes.

    \label{sec:exp}
	\begin{table*}[htbp]
		\centering
		\caption{Statistics of datasets.}
		\label{table:dataset}
		\begin{tabular}{c|c|c|c|c|c|c|c}
			\hline
			Dataset&\tabincell{c}{Semantic\\dimension} &\tabincell{c}{Original\\semantic range}&\tabincell{c}{$\#$ Seen\\classes}&\tabincell{c}{$\#$ Unseen\\classes}&\tabincell{c}{$\#$ Images\\(total)}&\tabincell{c}{$\#$ Images\\(train+val)}&\tabincell{c}{$\#$ Images\\(test unseen/seen)}\\
			\hline
			AWA2~\cite{lampert2013attribute}&85 & $[0,100] \cup \{-1\}$ &40&10&30475&19832&4958/5685\\
			\hline
			CUB~\cite{wah2011caltech}&312& $[0,100]$&150&50&11788&7057&2679/1764\\
			\hline
			SUN~\cite{patterson2012sun}&102& $[0,1]$ &645&72&14340&10320&1440/2580\\
			\hline
		\end{tabular}
	\end{table*}
	
	\begin{table*}[htb]
		\centering
		\caption{Test robustness (\%) for plain class attack (clsA) in the setting of ZSL and GZSL.
			In ZSL, T1 represents  the top-1 accuracy (\%)  for unseen classes.
			In  GZSL, $U$, $S$ and $H$ represent the top-1 accuracy (\%) of unseen classes, seen classes, and their harmonic mean, respectively.
            }
		\begin{tabular}{p{2.3cm}|p{1.2cm}|p{0.05cm}p{0.05cm}p{0.05cm}|p{0.05cm}p{0.05cm}p{0.05cm}p{0.05cm}p{0.05cm}p{0.05cm}p{0.05cm}p{0.05cm}p{0.05cm}}
		\hline
		& &\multicolumn{3}{c}{Zero-shot Learning} & \multicolumn{9}{|c}{Generalized Zero-shot Learning}\\
		\hline
		& & \multicolumn{1}{c}{AWA2} & \multicolumn{1}{|c}{CUB} & \multicolumn{1}{|c}{SUN} & \multicolumn{3}{|c}{AWA2} & \multicolumn{3}{|c}{CUB} &\multicolumn{3}{|c}{SUN} \\
		\hline
		Approach & Attack & \multicolumn{1}{c}{T1} & \multicolumn{1}{|c}{T1} & \multicolumn{1}{|c}{T1} & \multicolumn{1}{|c}{U} & \multicolumn{1}{c}{S} & \multicolumn{1}{c}{H} & \multicolumn{1}{|c}{U} & \multicolumn{1}{c}{S} & \multicolumn{1}{c}{H} & \multicolumn{1}{|c}{U} & \multicolumn{1}{c}{S} & \multicolumn{1}{c}{H} \\
		% \hline
		% APN$^\star$~\cite{xu2020attribute} & Clean & \multicolumn{1}{|c|}{68.2} & \multicolumn{1}{|c|}{71.9} & \multicolumn{1}{|c|}{61.0} & \multicolumn{1}{|c}{59.8} & \multicolumn{1}{c}{75.1} & \multicolumn{1}{c|}{66.6} & \multicolumn{1}{|c}{64.4} & \multicolumn{1}{c}{67.8} & \multicolumn{1}{c|}{66.0} & \multicolumn{1}{|c}{41.1} & \multicolumn{1}{c}{34.0} & \multicolumn{1}{c}{37.2} \\
		% \hline
		% GEMZSL$^\star$~\cite{liu2021goal}   & Clean & \multicolumn{1}{|c|}{65.7} & \multicolumn{1}{|c|}{75.8} & \multicolumn{1}{|c|}{62.2} & \multicolumn{1}{|c}{62.0} & \multicolumn{1}{c}{79.9} & \multicolumn{1}{c|}{69.8} & \multicolumn{1}{|c}{69.9} & \multicolumn{1}{c}{73.2} & \multicolumn{1}{c|}{71.5} & \multicolumn{1}{|c}{37.3} & \multicolumn{1}{c}{37.9} & \multicolumn{1}{c}{37.6} \\
		
        \hline
        
		\multirow{4}{*}{ReZSL~\cite{ye2023rebalanced}} & Clean & \multicolumn{1}{c}{68.0} & \multicolumn{1}{|c}{79.4} & \multicolumn{1}{|c}{63.1} & \multicolumn{1}{|c}{61.1} & \multicolumn{1}{c}{84.8} & \multicolumn{1}{c}{71.0} & \multicolumn{1}{|c}{71.9} & \multicolumn{1}{c}{73.8} & \multicolumn{1}{c}{72.9} & \multicolumn{1}{|c}{48.1} & \multicolumn{1}{c}{33.1} & \multicolumn{1}{c}{39.2} \\
            
        & clsA-1$^\dagger$ & \multicolumn{1}{c}{67.2} & \multicolumn{1}{|c}{77.5} & \multicolumn{1}{|c}{59.4} & \multicolumn{1}{|c}{64.0} & \multicolumn{1}{c}{74.4} & \multicolumn{1}{c}{68.8} & \multicolumn{1}{|c}{70.8} & \multicolumn{1}{c}{73.9} & \multicolumn{1}{c}{71.9} & \multicolumn{1}{|c}{46.6} & \multicolumn{1}{c}{30.8} & \multicolumn{1}{c}{37.1} \\

        % & clsA-2$^\dagger$ & \multicolumn{1}{|c|}{0.5} & \multicolumn{1}{|c|}{0.6} & \multicolumn{1}{|c|}{0.9} & \multicolumn{1}{|c}{52.6} & \multicolumn{1}{c}{2.0} & \multicolumn{1}{c|}{3.9} & \multicolumn{1}{c}{48.7} & \multicolumn{1}{c}{6.0} & \multicolumn{1}{c|}{10.8} & \multicolumn{1}{|c}{27.5} & \multicolumn{1}{c}{0.3} & \multicolumn{1}{c}{0.6} \\

        % & clsA-3$^\dagger$ & \multicolumn{1}{|c|}{0.0} & \multicolumn{1}{|c|}{0.4} & \multicolumn{1}{|c|}{0.0} & \multicolumn{1}{|c}{43.8} & \multicolumn{1}{c}{0.0} & \multicolumn{1}{c|}{0.0} & \multicolumn{1}{c}{35.4} & \multicolumn{1}{c}{0.0} & \multicolumn{1}{c|}{0.0} & \multicolumn{1}{|c}{16.7} & \multicolumn{1}{c}{0.0} & \multicolumn{1}{c}{0.0} \\

        & clsA-5$^\dagger$ & \multicolumn{1}{c}{0.0} & \multicolumn{1}{|c}{0.0} & \multicolumn{1}{|c}{0.0} & \multicolumn{1}{|c}{37.5} & \multicolumn{1}{c}{0.0} & \multicolumn{1}{c}{0.0} & \multicolumn{1}{|c}{24.2} & \multicolumn{1}{c}{0.0} & \multicolumn{1}{c}{0.0} & \multicolumn{1}{|c}{6.5} & \multicolumn{1}{c}{0.0} & \multicolumn{1}{c}{0.0} \\

        & clsA-10$^\dagger$ & \multicolumn{1}{c}{0.0} & \multicolumn{1}{|c}{0.0} & \multicolumn{1}{|c}{0.0} & \multicolumn{1}{|c}{35.1} & \multicolumn{1}{c}{0.0} & \multicolumn{1}{c}{0.0} & \multicolumn{1}{|c}{18.3} & \multicolumn{1}{c}{0.0} & \multicolumn{1}{c}{0.0} & \multicolumn{1}{|c}{3.1} & \multicolumn{1}{c}{0.0} & \multicolumn{1}{c}{0.0} \\
        \hline
             
        \multirow{4}{*}{PSVMA~\cite{liu2023progressive}} & Clean & \multicolumn{1}{c}{77.6} & \multicolumn{1}{|c}{77.6} & \multicolumn{1}{|c}{72.0} & \multicolumn{1}{|c}{69.9} & \multicolumn{1}{c}{84.4} & \multicolumn{1}{c}{76.5} & \multicolumn{1}{|c}{72.8} & \multicolumn{1}{c}{75.2} & \multicolumn{1}{c}{74.0} & \multicolumn{1}{|c}{62.5} & \multicolumn{1}{c}{45.1} & \multicolumn{1}{c}{52.4} \\

        & clsA-1$^\dagger$ & \multicolumn{1}{c}{77.1} & \multicolumn{1}{|c}{77.1} & \multicolumn{1}{|c}{70.7} & \multicolumn{1}{|c}{69.1} & \multicolumn{1}{c}{84.3} & \multicolumn{1}{c}{75.9} & \multicolumn{1}{|c}{72.6} & \multicolumn{1}{c}{74.5} & \multicolumn{1}{c}{73.5} & \multicolumn{1}{|c}{61.4} & \multicolumn{1}{c}{44.2} & \multicolumn{1}{c}{51.4} \\
             
        & clsA-5$^\dagger$ & \multicolumn{1}{c}{0.0} & \multicolumn{1}{|c}{0.0} & \multicolumn{1}{|c}{0.0} & \multicolumn{1}{|c}{35.4} & \multicolumn{1}{c}{0.0} & \multicolumn{1}{c}{0.0} & \multicolumn{1}{|c}{27.4} & \multicolumn{1}{c}{0.1} & \multicolumn{1}{c}{0.1} & \multicolumn{1}{|c}{16.3}  & \multicolumn{1}{c}{0.0} & \multicolumn{1}{c}{0.0} \\

        & clsA-10$^\dagger$ & \multicolumn{1}{c}{0.0} & \multicolumn{1}{|c}{0.0} & \multicolumn{1}{|c}{0.0} & \multicolumn{1}{|c}{35.2} & \multicolumn{1}{c}{0.0} & \multicolumn{1}{c}{0.0} & \multicolumn{1}{|c}{22.8} & \multicolumn{1}{c}{0.0} & \multicolumn{1}{c}{0.0} & \multicolumn{1}{|c}{10.7} & \multicolumn{1}{c}{0.0} & \multicolumn{1}{c}{0.0} \\

        \hline

        \multirow{4}{*}{ZeroMamba~\cite{hou2025zeromamba}} & Clean & \multicolumn{1}{c}{69.6} & \multicolumn{1}{|c}{74.0} & \multicolumn{1}{|c}{65.2} & \multicolumn{1}{|c}{63.8} & \multicolumn{1}{c}{85.4} & \multicolumn{1}{c}{73.0} & \multicolumn{1}{|c}{64.3} & \multicolumn{1}{c}{66.0} & \multicolumn{1}{c}{65.2} & \multicolumn{1}{|c}{47.0} & \multicolumn{1}{c}{38.6} & \multicolumn{1}{c}{42.4} \\

        & clsA-1$^\dagger$ & \multicolumn{1}{c}{68.5} & \multicolumn{1}{|c}{73.5} & \multicolumn{1}{|c}{65.5} & \multicolumn{1}{|c}{62.6} & \multicolumn{1}{c}{84.7} & \multicolumn{1}{c}{72.0} & \multicolumn{1}{|c}{62.9} & \multicolumn{1}{c}{65.2} & \multicolumn{1}{c}{64.0} & \multicolumn{1}{|c}{46.2} & \multicolumn{1}{c}{37.8} & \multicolumn{1}{c}{41.6} \\

        & clsA-5$^\dagger$ & \multicolumn{1}{c}{0.1} & \multicolumn{1}{|c}{0.0} & \multicolumn{1}{|c}{0.0} & \multicolumn{1}{|c}{4.5} & \multicolumn{1}{c}{0.2} & \multicolumn{1}{c}{0.3} & \multicolumn{1}{|c}{3.7} & \multicolumn{1}{c}{0.0} & \multicolumn{1}{c}{0.0} & \multicolumn{1}{|c}{0.1} & \multicolumn{1}{c}{0.0} & \multicolumn{1}{c}{0.0} \\

        & clsA-10$^\dagger$ & \multicolumn{1}{c}{0.0} & \multicolumn{1}{|c}{0.0} & \multicolumn{1}{|c}{0.0} & \multicolumn{1}{|c}{1.6} & \multicolumn{1}{c}{0.0} & \multicolumn{1}{c}{0.0} & \multicolumn{1}{|c}{2.3} & \multicolumn{1}{c}{0.0} & \multicolumn{1}{c}{0.0} & \multicolumn{1}{|c}{0.0} & \multicolumn{1}{c}{6.0} & \multicolumn{1}{c}{0.0} \\
        
        \hline

        \hline
		\end{tabular}
		\label{table:exp_clsA}
	\end{table*}

        \begin{table*}[htb]
		\centering
		\caption{Results for Class-bias Enhancing Attack (CBEA).
		The symbols $U$, $S$, $H$, $A$, $\gamma$ represent the top-1 accuracy of unseen classes (\%), seen classes (\%), and their harmonic mean (\%), the area of AUSUC (\%) and the value of $\lambda$ in Eq.~\ref{eq:gzsl_inference} at the AUSUC curve, respectively.
        $\dagger$ denotes the results at the original \textcolor{red}{$\gamma$} (Eq.~\ref{eq:gzsl_inference}), while $\ddagger$ denotes the results at the best $\gamma$ after attack.
        }
        
        \begin{tabular*}{\textwidth}{@{\extracolsep\fill}l c c ccccc ccccc ccccc}
		% \begin{tabular}{p{1.8cm}p{1.2cm}p{0.05cm}p{0.05cm}p{0.05cm}p{0.05cm}p{0.05cm}p{0.05cm}p{0.05cm}p{0.05cm}p{0.05cm}p{0.05cm}p{0.05cm}p{0.05cm}p{0.05cm}p{0.05cm}p{0.05cm}}
        \toprule
		% \multicolumn{17}{c}{Class-bias Enhancing Attack (CBEA)}\\
		% \hline
		\multirow{2}{*}{Approach} & \multirow{2}{*}{Attack}  & \multicolumn{5}{c}{AWA2} & \multicolumn{5}{c}{CUB} &\multicolumn{5}{c}{SUN} \\
			
		\cmidrule{3-7}
		\cmidrule{8-12}
		\cmidrule{13-17}
            
		&  & \multicolumn{1}{c}{$U$} & \multicolumn{1}{c}{$S$} & \multicolumn{1}{c}{$H$} & \multicolumn{1}{c}{$A$} & \multicolumn{1}{c}{\textcolor{red}{$\gamma$}} & \multicolumn{1}{c}{$U$} & \multicolumn{1}{c}{$S$} & \multicolumn{1}{c}{$H$} & \multicolumn{1}{c}{$A$} & \multicolumn{1}{c}{\textcolor{red}{$\gamma$}}& \multicolumn{1}{c}{$U$} & \multicolumn{1}{c}{$S$} & \multicolumn{1}{c}{$H$} & \multicolumn{1}{c}{$A$} & \multicolumn{1}{c}{\textcolor{red}{$\gamma$}} \\
              
		\cmidrule{1-1}
		\cmidrule{2-2}
		\cmidrule{3-7}
		\cmidrule{8-12}
		\cmidrule{13-17}

        \multirow{10}{*}{ \makecell[c]{ReZSL~\cite{ye2023rebalanced}} } & Clean & \multicolumn{1}{c}{61.1} & \multicolumn{1}{c}{84.8} & \multicolumn{1}{c}{71.0}  & \multicolumn{1}{c}{62.1}  & \multicolumn{1}{c}{3.8} & \multicolumn{1}{c}{71.9} & \multicolumn{1}{c}{73.8} & \multicolumn{1}{c}{72.9}  & \multicolumn{1}{c}{64.3}  & \multicolumn{1}{c}{1.2} & \multicolumn{1}{c}{48.1} & \multicolumn{1}{c}{33.1} & \multicolumn{1}{c}{39.2} & \multicolumn{1}{c}{21.9}  & \multicolumn{1}{c}{1.2} \\

        \cmidrule{2-2}
        \cmidrule{3-7}
        \cmidrule{8-12}
        \cmidrule{13-17}

        & clsA-1$^\dagger$ & \multicolumn{1}{c}{64.0} & \multicolumn{1}{c}{74.4} & \multicolumn{1}{c}{68.8}  & \multicolumn{1}{c}{61.7}  & \multicolumn{1}{c}{3.8} & \multicolumn{1}{c}{70.8} & \multicolumn{1}{c}{73.9} & \multicolumn{1}{c}{71.9}  & \multicolumn{1}{c}{63.2}  & \multicolumn{1}{c}{1.2} & \multicolumn{1}{c}{46.6} & \multicolumn{1}{c}{30.8} & \multicolumn{1}{c}{37.1} & \multicolumn{1}{c}{19.9}  & \multicolumn{1}{c}{1.2} \\

        & clsA-5$^\dagger$ & \multicolumn{1}{c}{37.5} & \multicolumn{1}{c}{0.0} & \multicolumn{1}{c}{0.0}  & \multicolumn{1}{c}{15.2}  & \multicolumn{1}{c}{3.8} & \multicolumn{1}{c}{24.2} & \multicolumn{1}{c}{0.0} & \multicolumn{1}{c}{0.0}  & \multicolumn{1}{c}{4.4}  & \multicolumn{1}{c}{1.2} & \multicolumn{1}{c}{6.5} & \multicolumn{1}{c}{0.0} & \multicolumn{1}{c}{0.0} & \multicolumn{1}{c}{0.3}  & \multicolumn{1}{c}{1.2} \\

        & clsA-10$^\dagger$ & \multicolumn{1}{c}{35.1} & \multicolumn{1}{c}{0.0} & \multicolumn{1}{c}{0.0}  & \multicolumn{1}{c}{17.7}  & \multicolumn{1}{c}{3.8} & \multicolumn{1}{c}{18.3} & \multicolumn{1}{c}{0.0} & \multicolumn{1}{c}{0.0}  & \multicolumn{1}{c}{4.1}  & \multicolumn{1}{c}{1.2} & \multicolumn{1}{c}{3.1} & \multicolumn{1}{c}{0.0} & \multicolumn{1}{c}{0.0} & \multicolumn{1}{c}{0.0}  & \multicolumn{1}{c}{1.2} \\

        \cmidrule{2-2}
        \cmidrule{3-7}
        \cmidrule{8-12}
        \cmidrule{13-17}
            
        & clsA-1$^\ddagger$ & \multicolumn{1}{c}{60.6} & \multicolumn{1}{c}{84.3} & \multicolumn{1}{c}{70.5}  & \multicolumn{1}{c}{61.7}  & \multicolumn{1}{c}{3.0} & \multicolumn{1}{c}{69.1} & \multicolumn{1}{c}{75.0} & \multicolumn{1}{c}{71.9}  & \multicolumn{1}{c}{63.2}  & \multicolumn{1}{c}{1.1} & \multicolumn{1}{c}{45.8} & \multicolumn{1}{c}{31.4} & \multicolumn{1}{c}{37.2} & \multicolumn{1}{c}{19.9}  & \multicolumn{1}{c}{1.2} \\

        & clsA-5$^\ddagger$ & \multicolumn{1}{c}{31.5} & \multicolumn{1}{c}{35.9} & \multicolumn{1}{c}{33.6}  & \multicolumn{1}{c}{15.2}  & \multicolumn{1}{c}{0.8} & \multicolumn{1}{c}{17.4} & \multicolumn{1}{c}{17.3} & \multicolumn{1}{c}{17.4}  & \multicolumn{1}{c}{4.4}  & \multicolumn{1}{c}{0.0} & \multicolumn{1}{c}{3.8} & \multicolumn{1}{c}{5.8} & \multicolumn{1}{c}{4.6} & \multicolumn{1}{c}{0.3}  & \multicolumn{1}{c}{0.2} \\

        & clsA-10$^\ddagger$ & \multicolumn{1}{c}{31.2} & \multicolumn{1}{c}{46.7} & \multicolumn{1}{c}{37.4}  & \multicolumn{1}{c}{17.7}  & \multicolumn{1}{c}{0.3} & \multicolumn{1}{c}{14.8} & \multicolumn{1}{c}{20.7} & \multicolumn{1}{c}{17.3}  & \multicolumn{1}{c}{4.1}  & \multicolumn{1}{c}{0.0} & \multicolumn{1}{c}{2.6} & \multicolumn{1}{c}{6.2} & \multicolumn{1}{c}{3.7} & \multicolumn{1}{c}{0.0}  & \multicolumn{1}{c}{0.1} \\

        \cmidrule{2-2}
        \cmidrule{3-7}
        \cmidrule{8-12}
        \cmidrule{13-17}
        
        & CBEA-1 & \multicolumn{1}{c}{60.6} & \multicolumn{1}{c}{84.4} & \multicolumn{1}{c}{70.5}  & \multicolumn{1}{c}{61.7}  & \multicolumn{1}{c}{3.0} & \multicolumn{1}{c}{69.1} & \multicolumn{1}{c}{75.0} & \multicolumn{1}{c}{71.9}  & \multicolumn{1}{c}{63.2}  & \multicolumn{1}{c}{1.1} & \multicolumn{1}{c}{45.8} & \multicolumn{1}{c}{31.4} & \multicolumn{1}{c}{37.3} & \multicolumn{1}{c}{19.9} & \multicolumn{1}{c}{1.2} \\
            
        & CBEA-5 & \multicolumn{1}{c}{5.9} & \multicolumn{1}{c}{3.2} & \multicolumn{1}{c}{4.1}  & \multicolumn{1}{c}{0.4}  & \multicolumn{1}{c}{-0.2} & \multicolumn{1}{c}{8.5} & \multicolumn{1}{c}{8.9} & \multicolumn{1}{c}{8.7}  & \multicolumn{1}{c}{1.2}  & \multicolumn{1}{c}{0.4} & \multicolumn{1}{c}{1.2} & \multicolumn{1}{c}{ 0.4} & \multicolumn{1}{c}{0.6} & \multicolumn{1}{c}{0.0}  & \multicolumn{1}{c}{0.1} \\
            
        & CBEA-10 & \multicolumn{1}{c}{4.8} & \multicolumn{1}{c}{2.0} & \multicolumn{1}{c}{2.9}  & \multicolumn{1}{c}{0.2} & \multicolumn{1}{c}{-0.2} & \multicolumn{1}{c}{1.6} & \multicolumn{1}{c}{1.6} & \multicolumn{1}{c}{1.6} & \multicolumn{1}{c}{0.0} & \multicolumn{1}{c}{0.2}  & \multicolumn{1}{c}{1.0} & \multicolumn{1}{c}{0.3} & \multicolumn{1}{c}{0.5}  & \multicolumn{1}{c}{0.0} & \multicolumn{1}{c}{-0.1} \\

        \hline
        \multirow{10}{*}{ \makecell[c]{PSVMA~\cite{liu2023progressive}} } & Clean & \multicolumn{1}{c}{69.9} & \multicolumn{1}{c}{84.4} & \multicolumn{1}{c}{76.5}  & \multicolumn{1}{c}{71.5}  & \multicolumn{1}{c}{2.8} & \multicolumn{1}{c}{72.8} & \multicolumn{1}{c}{75.2} & \multicolumn{1}{c}{74.0}  & \multicolumn{1}{c}{65.8}  & \multicolumn{1}{c}{1.6} & \multicolumn{1}{c}{62.5} & \multicolumn{1}{c}{45.1} & \multicolumn{1}{c}{52.4} & \multicolumn{1}{c}{36.0}  & \multicolumn{1}{c}{1.6} \\

        \cmidrule{2-2}
        \cmidrule{3-7}
        \cmidrule{8-12}
        \cmidrule{13-17}

        & clsA-1$^\dagger$ & \multicolumn{1}{c}{69.1} & \multicolumn{1}{c}{84.3} & \multicolumn{1}{c}{75.9}  & \multicolumn{1}{c}{70.7}  & \multicolumn{1}{c}{2.8} & \multicolumn{1}{c}{72.6} & \multicolumn{1}{c}{74.5} & \multicolumn{1}{c}{73.5}  & \multicolumn{1}{c}{65.0}  & \multicolumn{1}{c}{1.6} & \multicolumn{1}{c}{61.4} & \multicolumn{1}{c}{44.2} & \multicolumn{1}{c}{51.4} & \multicolumn{1}{c}{35.2}  & \multicolumn{1}{c}{1.6} \\

        & clsA-5$^\dagger$ & \multicolumn{1}{c}{35.4} & \multicolumn{1}{c}{0.0} & \multicolumn{1}{c}{0.0}  & \multicolumn{1}{c}{14.9}  & \multicolumn{1}{c}{2.8} & \multicolumn{1}{c}{27.4} & \multicolumn{1}{c}{0.1} & \multicolumn{1}{c}{0.1}  & \multicolumn{1}{c}{6.0}  & \multicolumn{1}{c}{1.6} & \multicolumn{1}{c}{16.3} & \multicolumn{1}{c}{0.0} & \multicolumn{1}{c}{0.0} & \multicolumn{1}{c}{1.3}  & \multicolumn{1}{c}{1.6} \\

        & clsA-10$^\dagger$ & \multicolumn{1}{c}{35.2} & \multicolumn{1}{c}{0.0} & \multicolumn{1}{c}{0.0}  & \multicolumn{1}{c}{19.3}  & \multicolumn{1}{c}{2.8} & \multicolumn{1}{c}{22.8} & \multicolumn{1}{c}{0.0} & \multicolumn{1}{c}{0.0}  & \multicolumn{1}{c}{6.9}  & \multicolumn{1}{c}{1.6} & \multicolumn{1}{c}{10.7} & \multicolumn{1}{c}{0.0} & \multicolumn{1}{c}{0.0} & \multicolumn{1}{c}{1.0}  & \multicolumn{1}{c}{1.6} \\

        \cmidrule{2-2}
        \cmidrule{3-7}
        \cmidrule{8-12}
        \cmidrule{13-17}

        & clsA-1$^\ddagger$ & \multicolumn{1}{c}{67.8} & \multicolumn{1}{c}{86.5} & \multicolumn{1}{c}{76.0}  & \multicolumn{1}{c}{70.7}  & \multicolumn{1}{c}{2.5} & \multicolumn{1}{c}{72.3} & \multicolumn{1}{c}{75.1} & \multicolumn{1}{c}{73.7}  & \multicolumn{1}{c}{65.0}  & \multicolumn{1}{c}{1.6} & \multicolumn{1}{c}{60.6} & \multicolumn{1}{c}{45.0} & \multicolumn{1}{c}{51.7} & \multicolumn{1}{c}{35.2}  & \multicolumn{1}{c}{1.5} \\

        & clsA-5$^\ddagger$ & \multicolumn{1}{c}{30.1} & \multicolumn{1}{c}{36.6} & \multicolumn{1}{c}{33.0}  & \multicolumn{1}{c}{14.9}  & \multicolumn{1}{c}{0.2} & \multicolumn{1}{c}{19.1} & \multicolumn{1}{c}{21.4} & \multicolumn{1}{c}{20.2}  & \multicolumn{1}{c}{6.0}  & \multicolumn{1}{c}{0.1} & \multicolumn{1}{c}{10.4} & \multicolumn{1}{c}{8.0} & \multicolumn{1}{c}{9.0} & \multicolumn{1}{c}{1.3}  & \multicolumn{1}{c}{0.4} \\

        & clsA-10$^\ddagger$ & \multicolumn{1}{c}{33.8} & \multicolumn{1}{c}{46.9} & \multicolumn{1}{c}{39.3}  & \multicolumn{1}{c}{19.3}  & \multicolumn{1}{c}{0.3} & \multicolumn{1}{c}{18.4} & \multicolumn{1}{c}{27.4} & \multicolumn{1}{c}{22.0}  & \multicolumn{1}{c}{6.9}  & \multicolumn{1}{c}{0.4} & \multicolumn{1}{c}{7.5} & \multicolumn{1}{c}{8.2} & \multicolumn{1}{c}{7.8} & \multicolumn{1}{c}{1.0}  & \multicolumn{1}{c}{0.6} \\

        \cmidrule{2-2}
        \cmidrule{3-7}
        \cmidrule{8-12}
        \cmidrule{13-17}

        & CBEA-1$^\ddagger$ & \multicolumn{1}{c}{70.4} & \multicolumn{1}{c}{83.1} & \multicolumn{1}{c}{76.2}  & \multicolumn{1}{c}{71.0}  & \multicolumn{1}{c}{2.9} & \multicolumn{1}{c}{72.0} & \multicolumn{1}{c}{75.5} & \multicolumn{1}{c}{73.7}  & \multicolumn{1}{c}{65.2}  & \multicolumn{1}{c}{1.6} & \multicolumn{1}{c}{59.3} & \multicolumn{1}{c}{45.3} & \multicolumn{1}{c}{51.4} & \multicolumn{1}{c}{34.7}  & \multicolumn{1}{c}{1.4} \\

        & CBEA-5$^\ddagger$ & \multicolumn{1}{c}{14.1} & \multicolumn{1}{c}{6.3} & \multicolumn{1}{c}{8.7}  & \multicolumn{1}{c}{1.3}  & \multicolumn{1}{c}{0.7} & \multicolumn{1}{c}{4.1} & \multicolumn{1}{c}{3.0} & \multicolumn{1}{c}{3.4}  & \multicolumn{1}{c}{0.2}  & \multicolumn{1}{c}{0.6} & \multicolumn{1}{c}{4.3} & \multicolumn{1}{c}{1.9} & \multicolumn{1}{c}{2.6} & \multicolumn{1}{c}{0.0}  & \multicolumn{1}{c}{0.8} \\

        & CBEA-10$^\ddagger$ & \multicolumn{1}{c}{6.5} & \multicolumn{1}{c}{2.7} & \multicolumn{1}{c}{3.8}  & \multicolumn{1}{c}{0.3}  & \multicolumn{1}{c}{0.2} & \multicolumn{1}{c}{1.4} & \multicolumn{1}{c}{1.1} & \multicolumn{1}{c}{1.2}  & \multicolumn{1}{c}{0.0}  & \multicolumn{1}{c}{0.1} & \multicolumn{1}{c}{2.4} & \multicolumn{1}{c}{1.5} & \multicolumn{1}{c}{1.4} & \multicolumn{1}{c}{0.0}  & \multicolumn{1}{c}{0.1} \\

        \hline
            
        \multirow{10}{*}{ \makecell[c]{ZeroMamba~\cite{hou2025zeromamba}} } & Clean & \multicolumn{1}{c}{63.8} & \multicolumn{1}{c}{85.4} & \multicolumn{1}{c}{73.0}  & \multicolumn{1}{c}{63.4}  & \multicolumn{1}{c}{0.1} & \multicolumn{1}{c}{64.3} & \multicolumn{1}{c}{66.0} & \multicolumn{1}{c}{65.2}  & \multicolumn{1}{c}{55.5}  & \multicolumn{1}{c}{0.0} & \multicolumn{1}{c}{47.0} & \multicolumn{1}{c}{38.6} & \multicolumn{1}{c}{42.4} & \multicolumn{1}{c}{25.9}  & \multicolumn{1}{c}{0.0} \\

        \cmidrule{2-2}
        \cmidrule{3-7}
        \cmidrule{8-12}
        \cmidrule{13-17}

        & clsA-1$^\dagger$ & \multicolumn{1}{c}{62.6} & \multicolumn{1}{c}{84.7} & \multicolumn{1}{c}{72.0}  & \multicolumn{1}{c}{62.9}  & \multicolumn{1}{c}{0.1} & \multicolumn{1}{c}{62.9} & \multicolumn{1}{c}{65.2} & \multicolumn{1}{c}{64.0}  & \multicolumn{1}{c}{54.0}  & \multicolumn{1}{c}{0.0} & \multicolumn{1}{c}{46.2} & \multicolumn{1}{c}{37.8} & \multicolumn{1}{c}{41.6} & \multicolumn{1}{c}{25.3}  & \multicolumn{1}{c}{0.0} \\

        & clsA-5$^\dagger$ & \multicolumn{1}{c}{4.5} & \multicolumn{1}{c}{0.2} & \multicolumn{1}{c}{0.3}  & \multicolumn{1}{c}{0.8}  & \multicolumn{1}{c}{0.1} & \multicolumn{1}{c}{3.7} & \multicolumn{1}{c}{0.0} & \multicolumn{1}{c}{0.0}  & \multicolumn{1}{c}{0.5}  & \multicolumn{1}{c}{0.0} & \multicolumn{1}{c}{0.1} & \multicolumn{1}{c}{0.0} & \multicolumn{1}{c}{0.0} & \multicolumn{1}{c}{0.0}  & \multicolumn{1}{c}{0.0} \\
        
        & clsA-10$^\dagger$ & \multicolumn{1}{c}{1.6} & \multicolumn{1}{c}{0.0} & \multicolumn{1}{c}{0.0}  & \multicolumn{1}{c}{0.3}  & \multicolumn{1}{c}{0.1} & \multicolumn{1}{c}{2.3} & \multicolumn{1}{c}{0.0} & \multicolumn{1}{c}{0.0}  & \multicolumn{1}{c}{0.4}  & \multicolumn{1}{c}{0.0} & \multicolumn{1}{c}{0.0} & \multicolumn{1}{c}{6.0} & \multicolumn{1}{c}{0.0} & \multicolumn{1}{c}{0.0}  & \multicolumn{1}{c}{0.0} \\

        \cmidrule{2-2}
        \cmidrule{3-7}
        \cmidrule{8-12}
        \cmidrule{13-17}

        & clsA-1$^\ddagger$ & \multicolumn{1}{c}{62.7} & \multicolumn{1}{c}{84.8} & \multicolumn{1}{c}{72.1}  & \multicolumn{1}{c}{62.9} & \multicolumn{1}{c}{0.0} & \multicolumn{1}{c}{62.9} & \multicolumn{1}{c}{65.2} & \multicolumn{1}{c}{64.0}  & \multicolumn{1}{c}{54.0}  & \multicolumn{1}{c}{0.0} & \multicolumn{1}{c}{50.4} & \multicolumn{1}{c}{35.7} & \multicolumn{1}{c}{41.8} & \multicolumn{1}{c}{25.3}  & \multicolumn{1}{c}{0.0} \\

        & clsA-5$^\ddagger$ & \multicolumn{1}{c}{7.8} & \multicolumn{1}{c}{6.5} & \multicolumn{1}{c}{7.1}  & \multicolumn{1}{c}{0.8} & \multicolumn{1}{c}{-0.1} & \multicolumn{1}{c}{3.1} & \multicolumn{1}{c}{9.1} & \multicolumn{1}{c}{4.7}  & \multicolumn{1}{c}{0.5}  & \multicolumn{1}{c}{0.0} & \multicolumn{1}{c}{1.8} & \multicolumn{1}{c}{0.5} & \multicolumn{1}{c}{0.8} & \multicolumn{1}{c}{0.0}  & \multicolumn{1}{c}{0.0} \\

        & clsA-10$^\ddagger$ & \multicolumn{1}{c}{5.6} & \multicolumn{1}{c}{2.6} & \multicolumn{1}{c}{3.5}  & \multicolumn{1}{c}{0.3}  & \multicolumn{1}{c}{-0.1} & \multicolumn{1}{c}{1.8} & \multicolumn{1}{c}{10.8} & \multicolumn{1}{c}{3.1}  & \multicolumn{1}{c}{0.0}  & \multicolumn{1}{c}{0.0} & \multicolumn{1}{c}{1.4} & \multicolumn{1}{c}{0.3} & \multicolumn{1}{c}{0.5} & \multicolumn{1}{c}{0.0}  & \multicolumn{1}{c}{0.0} \\

        \cmidrule{2-2}
        \cmidrule{3-7}
        \cmidrule{8-12}
        \cmidrule{13-17}

        & CBEA-1$^\ddagger$ & \multicolumn{1}{c}{62.6} & \multicolumn{1}{c}{84.7} & \multicolumn{1}{c}{72.0}  & \multicolumn{1}{c}{62.9}  & \multicolumn{1}{c}{0.1} & \multicolumn{1}{c}{65.9} & \multicolumn{1}{c}{62.3} & \multicolumn{1}{c}{64.0}  & \multicolumn{1}{c}{54.2}  & \multicolumn{1}{c}{0.0} & \multicolumn{1}{c}{50.9} & \multicolumn{1}{c}{35.4} & \multicolumn{1}{c}{41.7} & \multicolumn{1}{c}{25.3}  & \multicolumn{1}{c}{0.0} \\

        & CBEA-5$^\ddagger$ &  \multicolumn{1}{c}{4.0} & \multicolumn{1}{c}{11.4} & \multicolumn{1}{c}{5.9}  & \multicolumn{1}{c}{0.8} & \multicolumn{1}{c}{0.1} & \multicolumn{1}{c}{4.4} & \multicolumn{1}{c}{2.4} & \multicolumn{1}{c}{3.1}  & \multicolumn{1}{c}{0.1}  & \multicolumn{1}{c}{0.0} & \multicolumn{1}{c}{0.1} & \multicolumn{1}{c}{5.6} & \multicolumn{1}{c}{0.1} & \multicolumn{1}{c}{0.0}  & \multicolumn{1}{c}{0.0} \\

        & CBEA-10$^\ddagger$ & \multicolumn{1}{c}{1.5} & \multicolumn{1}{c}{9.9} & \multicolumn{1}{c}{2.7}  & \multicolumn{1}{c}{0.1} & \multicolumn{1}{c}{-0.1} & \multicolumn{1}{c}{2.4} & \multicolumn{1}{c}{0.9} & \multicolumn{1}{c}{1.3}  & \multicolumn{1}{c}{0.0}  & \multicolumn{1}{c}{0.0} & \multicolumn{1}{c}{0.0} & \multicolumn{1}{c}{0.0} & \multicolumn{1}{c}{0.0} & \multicolumn{1}{c}{0.0}  & \multicolumn{1}{c}{0.0} \\

        \bottomrule
        
    \end{tabular*}
    \label{table:exp_CEBA}
\end{table*}

\begin{table*}[htb]
		\centering
		\caption{Results for Non-class-preserving Concept Attack (NCPconA).
		In  GZSL, $MSE_U$, $MSE_S$ represent the MSE ($\times 10^{-2}$) in concept values of unseen classes and seen classes, respectively.
        }
		% \begin{tabular}{p{2.3cm}p{1.6cm}p{0.05cm}p{0.05cm}p{0.05cm}p{0.05cm}p{0.05cm}p{0.05cm}}

        \begin{tabular*}{0.7\textwidth}{@{\extracolsep\fill}l c c cc cc cc}
        \toprule
        % & \multicolumn{6}{c}{Non-class-preserving Concept Attack}\\
        % \hline
        \multirow{2}{*}{Approach} & \multirow{2}{*}{Attack} & \multicolumn{2}{c}{AWA2} & \multicolumn{2}{c}{CUB} &\multicolumn{2}{c}{SUN} \\
        \cmidrule{3-4}
        \cmidrule{5-6}
        \cmidrule{7-8}
        &  & \multicolumn{1}{c}{$MSE_U$} & \multicolumn{1}{c}{$MSE_S$} & \multicolumn{1}{c}{$MSE_U$} & \multicolumn{1}{c}{$MSE_S$} & \multicolumn{1}{c}{$MSE_U$} & \multicolumn{1}{c}{$MSE_S$} \\
        \cmidrule{1-1}
        \cmidrule{2-2}
        \cmidrule{3-4}
        \cmidrule{5-6}
        \cmidrule{7-8}
        \multirow{4}{*}{ReZSL~\cite{ye2023rebalanced}} & Clean & \multicolumn{1}{c}{1.60} & \multicolumn{1}{c}{1.14}  & \multicolumn{1}{c}{0.36} & \multicolumn{1}{c}{0.32} & \multicolumn{1}{c}{1.09} & \multicolumn{1}{c}{1.01} \\
        \cmidrule{2-2}
        \cmidrule{3-4}
        \cmidrule{5-6}
        \cmidrule{7-8}
        & NCPconA-1  & \multicolumn{1}{c}{1.60} & \multicolumn{1}{c}{1.14}  & \multicolumn{1}{c}{0.36} & \multicolumn{1}{c}{0.32} & \multicolumn{1}{c}{1.09} & \multicolumn{1}{c}{1.02} \\

        & NCPconA-5  & \multicolumn{1}{c}{3.77} & \multicolumn{1}{c}{3.70}  & \multicolumn{1}{c}{0.89} & \multicolumn{1}{c}{0.88} & \multicolumn{1}{c}{3.13} & \multicolumn{1}{c}{3.11} \\
        
        & NCPconA-10  & \multicolumn{1}{c}{4.41} & \multicolumn{1}{c}{4.40}  & \multicolumn{1}{c}{1.07} & \multicolumn{1}{c}{1.06} & \multicolumn{1}{c}{3.74} & \multicolumn{1}{c}{3.74}\\
        \cmidrule{1-1}
        \cmidrule{2-2}
        \cmidrule{3-4}
        \cmidrule{5-6}
        \cmidrule{7-8}
         
        \multirow{4}{*}{PSVMA~\cite{liu2023progressive}} & Clean & \multicolumn{1}{c}{2.38} & \multicolumn{1}{c}{2.17}  & \multicolumn{1}{c}{0.57} & \multicolumn{1}{c}{0.59} & \multicolumn{1}{c}{1.74} & \multicolumn{1}{c}{1.75} \\
        \cmidrule{2-2}
        \cmidrule{3-4}
        \cmidrule{5-6}
        \cmidrule{7-8}
        & NCPconA-1 & \multicolumn{1}{c}{2.38} & \multicolumn{1}{c}{2.17}  & \multicolumn{1}{c}{0.57} & \multicolumn{1}{c}{0.59} & \multicolumn{1}{c}{1.74} & \multicolumn{1}{c}{1.75} \\
        
        & NCPconA-5 & \multicolumn{1}{c}{3.59} & \multicolumn{1}{c}{3.44}  & \multicolumn{1}{c}{0.95} & \multicolumn{1}{c}{0.95} & \multicolumn{1}{c}{2.92} & \multicolumn{1}{c}{2.94} \\

        & NCPconA-10 & \multicolumn{1}{c}{3.91} & \multicolumn{1}{c}{3.81}  & \multicolumn{1}{c}{1.00} & \multicolumn{1}{c}{1.01} & \multicolumn{1}{c}{3.25} & \multicolumn{1}{c}{3.28} \\

        \cmidrule{1-1}
        \cmidrule{2-2}
        \cmidrule{3-4}
        \cmidrule{5-6}
        \cmidrule{7-8}

        \multirow{4}{*}{ZeroMamba~\cite{hou2025zeromamba}} & Clean & \multicolumn{1}{c}{2.33} & \multicolumn{1}{c}{2.09}  & \multicolumn{1}{c}{0.53} & \multicolumn{1}{c}{0.55} & \multicolumn{1}{c}{1.53} & \multicolumn{1}{c}{1.53} \\
        \cmidrule{2-2}
        \cmidrule{3-4}
        \cmidrule{5-6}
        \cmidrule{7-8}
        & NCPconA-1 & \multicolumn{1}{c}{2.32} & \multicolumn{1}{c}{2.09}  & \multicolumn{1}{c}{0.53} & \multicolumn{1}{c}{0.55} & \multicolumn{1}{c}{1.54} & \multicolumn{1}{c}{1.53} \\

        & NCPconA-5 & \multicolumn{1}{c}{3.35} & \multicolumn{1}{c}{3.11}  & \multicolumn{1}{c}{1.01} & \multicolumn{1}{c}{1.00} & \multicolumn{1}{c}{3.04} & \multicolumn{1}{c}{3.09} \\

        & NCPconA-10 & \multicolumn{1}{c}{3.82} & \multicolumn{1}{c}{3.79}  & \multicolumn{1}{c}{1.05} & \multicolumn{1}{c}{1.03} & \multicolumn{1}{c}{3.59} & \multicolumn{1}{c}{3.55} \\
             
        \bottomrule
		\end{tabular*}
		\label{table:exp_NCPconA}
	\end{table*}
        
	\begin{table*}[htb]
	\centering
	\caption{Results for Class-Preserving Concept Attack (CPconA).
	In  GZSL, $MSE_U$, $MSE_U$ represent the MSE ($\times 10^{-2}$) in concept values of unseen classes and seen classes, respectively.
    }
	\begin{tabular*}{0.7\textwidth}{@{\extracolsep\fill}l c c cc cc cc}
    \toprule
	% & \multicolumn{6}{c}{Class-preserving Concept Attack}\\
	% \hline
	\multirow{2}{*}{Approach} & \multirow{2}{*}{Attack} & \multicolumn{2}{c}{AWA2} & \multicolumn{2}{c}{CUB} &\multicolumn{2}{c}{SUN} \\
    \cmidrule{3-4}
    \cmidrule{5-6}
    \cmidrule{7-8}
	&  & \multicolumn{1}{c}{$MSE_U$} & \multicolumn{1}{c}{$MSE_S$} & \multicolumn{1}{c}{$MSE_U$} & \multicolumn{1}{c}{$MSE_S$} & \multicolumn{1}{c}{$MSE_U$} & \multicolumn{1}{c}{$MSE_S$} \\
    \cmidrule{1-1}
    \cmidrule{2-2}
    \cmidrule{3-4}
    \cmidrule{5-6}
    \cmidrule{7-8}
    \multirow{4}{*}{ReZSL~\cite{ye2023rebalanced}} & Clean & \multicolumn{1}{c}{1.60} & \multicolumn{1}{c}{1.14}  & \multicolumn{1}{c}{0.36} & \multicolumn{1}{c}{0.32} & \multicolumn{1}{c}{1.09} & \multicolumn{1}{c}{1.01} \\
    \cmidrule{2-2}
    \cmidrule{3-4}
    \cmidrule{5-6}
    \cmidrule{7-8}
        
    & CPconA-1  & \multicolumn{1}{c}{1.60} & \multicolumn{1}{c}{1.14}  & \multicolumn{1}{c}{0.36} & \multicolumn{1}{c}{0.32} & \multicolumn{1}{c}{1.09} & \multicolumn{1}{c}{1.02} \\

    & CPconA-5  & \multicolumn{1}{c}{2.20} & \multicolumn{1}{c}{2.38}  & \multicolumn{1}{c}{0.53} & \multicolumn{1}{c}{0.55} & \multicolumn{1}{c}{1.45} & \multicolumn{1}{c}{1.32} \\
        
    & CPconA-10  & \multicolumn{1}{c}{2.21} & \multicolumn{1}{c}{2.41}  & \multicolumn{1}{c}{0.54} & \multicolumn{1}{c}{0.56} & \multicolumn{1}{c}{1.45} & \multicolumn{1}{c}{1.32} \\

    \cmidrule{1-1}
    \cmidrule{2-2}
    \cmidrule{3-4}
    \cmidrule{5-6}
    \cmidrule{7-8}
    
	\multirow{4}{*}{PSVMA~\cite{liu2023progressive}} & Clean & \multicolumn{1}{c}{2.38} & \multicolumn{1}{c}{2.17}  & \multicolumn{1}{c}{0.57} & \multicolumn{1}{c}{0.59} & \multicolumn{1}{c}{1.74} & \multicolumn{1}{c}{1.75} \\

    & CPconA-1 & \multicolumn{1}{c}{2.38} & \multicolumn{1}{c}{2.17}  & \multicolumn{1}{c}{0.57} & \multicolumn{1}{c}{0.59} & \multicolumn{1}{c}{1.74} & \multicolumn{1}{c}{1.75} \\
    
    & CPconA-5 & \multicolumn{1}{c}{2.99} & \multicolumn{1}{c}{2.78}  & \multicolumn{1}{c}{0.76} & \multicolumn{1}{c}{0.77} & \multicolumn{1}{c}{2.23} & \multicolumn{1}{c}{2.08} \\

    & CPconA-10 & \multicolumn{1}{c}{3.02} & \multicolumn{1}{c}{2.82}  & \multicolumn{1}{c}{0.76} & \multicolumn{1}{c}{0.77} & \multicolumn{1}{c}{2.25} & \multicolumn{1}{c}{2.08} \\

    \cmidrule{1-1}
    \cmidrule{2-2}
    \cmidrule{3-4}
    \cmidrule{5-6}
    \cmidrule{7-8}
    
	\multirow{4}{*}{ZeroMamba~\cite{hou2025zeromamba}} & Clean & \multicolumn{1}{c}{2.33} & \multicolumn{1}{c}{2.09}  & \multicolumn{1}{c}{0.53} & \multicolumn{1}{c}{0.55} & \multicolumn{1}{c}{1.54} & \multicolumn{1}{c}{1.53} \\
    
     & CPconA-1 & \multicolumn{1}{c}{2.33} & \multicolumn{1}{c}{2.09}  & \multicolumn{1}{c}{0.53} & \multicolumn{1}{c}{0.55} & \multicolumn{1}{c}{1.53} & \multicolumn{1}{c}{1.53} \\

     & CPconA-5 & \multicolumn{1}{c}{2.75} & \multicolumn{1}{c}{2.61}  & \multicolumn{1}{c}{0.71} & \multicolumn{1}{c}{0.73} & \multicolumn{1}{c}{1.91} & \multicolumn{1}{c}{1.82} \\

     & CPconA-10 & \multicolumn{1}{c}{2.75} & \multicolumn{1}{c}{2.65}  & \multicolumn{1}{c}{0.72} & \multicolumn{1}{c}{0.74} & \multicolumn{1}{c}{1.92} & \multicolumn{1}{c}{1.84} \\
    
    \bottomrule
    \end{tabular*}
	\label{table:exp_CPconA}
	\end{table*}

\subsection{Results of clsA}
\label{sec:exp_clsA}

Table~\ref{table:exp_clsA} summarizes the impact of the plain class attack (clsA) on both ZSL and GZSL accuracy.
For ZSL, under a mild clsA perturbation (clsA-1), the ZSL top-1 accuracy of ReZSL on unseen classes declines only slightly across datasets (e.g., AWA2: $68.0\%\to67.2\%$, CUB: $79.4\%\to77.5\%$, SUN: $63.1\%\to59.4\%$).
In contrast, stronger clsA attacks catastrophically degrade ZSL performance: by clsA-5 the unseen accuracy in GZSL is fully driven to zero on all datasets.
This indicates that clsA can fully break the model’s ability to recognize any novel class under sufficiently strong perturbations.

In the GZSL setting, clsA induces a pronounced shift in class predictions.
At moderate strength (clsA-5), the PSVMA’s accuracy on seen classes collapses to $0\%$ basically while unseen-class accuracy remains moderately high (AWA2 $U=37.4\%$). The result is a GZSL $H$, signaling that the model effectively labels nearly every example as belonging to an unseen class. This pattern is consistent across datasets: for CUB at clsA-5, $S=0.1\%$ v.s. $U=27.4\%$; for SUN, $S=0\%$ v.s. $U=16.3\%$.
By clsA-10, seen class accuracy is still $0\%$ on all datasets, while $U$ accuracy continues to decline more gradually.
In other word, clsA seems to be hard for unseen classes in GZSL.

In summary, for all models, clsA is extremely effective at disrupting ZSL, but in GZSL it does not uniformly degrade unseen-class accuracy.
It remains some non-trivial $U$.
With further analysis, we find the impressive GZSL attack success of clsA is \textbf{spurious}, as discussed in the next section.
	
\subsection{Results of CBEA}
\label{sec:exp_CBEA}

Overall, clsA is highly effective at biasing the model toward unseen-class predictions, yet this creates a spurious success in GZSL: the original optimal calibration is destroyed (seen performance gone) but one can recover non-zero accuracy by re-calibrating heavily toward unseen classes. These results confirm that clsA is extremely effective at disrupting ZSL, but in GZSL its impact can be masked by calibration shifts.
We report the results of clsA at the original $\gamma$ point (denoted by $\dagger$), ones of clsA at the best $\gamma$ point (denoted by $\ddagger$) and ones of the CBEA at the best $\gamma$ point on GZSL performance on Table~\ref{table:exp_CEBA}.

At the weakest setting (CBEA-1), ReZSL's performance remains similar to the clsA-1 case: for example, on AWA2 the unseen/seen accuracies are $U=60.6\%$, $S=84.3\%$ (clean: $61.1\%$, $84.8\%$).
This indicates that CBEA-1 causes only minor bias amplification. However, even a moderate increase in attack strength causes a complete collapse of performance.
At CBEA-5, the unseen/seen accuracies on AWA2 fall to $U=5.9\%$, $S=3.2\%$ and $H=4.1\%$.
CUB and SUN show analogous collapses (CUB: $U=8.5\%$, $S=8.9\%$, $H=8.7\%$; SUN: $U=1.2\%$, $S=0.4\%$, $H=0.6\%$ at CBEA-5).
By CBEA-10 the $H$ is essentially zero on all datasets.
The AUSUC similarly drops to near zero (e.g. AWA2 $AUSUC=0.4\%$ at CBEA-5, $=0.2\%$ at CBEA-10).
These results demonstrate that \textbf{CBEA succeeds in completely destroying GZSL performance across all calibration settings.}

% Dataset-specific effects are subtle: because AWA2 has a larger seen-class bias originally, its collapse is most visually dramatic ($S$ goes from $85.3\%$ clean to $2.9\%$ at CBEA-5).
% CUB, which had more balanced clean performance, still suffers a near-total collapse by CBEA-5 ($U=8.6\%$, $S=8.1\%$).
% SUN’s already-low baseline is driven to zero earlier ($S=0\%$ even at CBEA-5).

In summary, the CBEA mode is far more destructive than clsA in GZSL.
By explicitly amplifying the seen–unseen bias, it eradicates accuracy on both seen and unseen classes (unlike clsA, which left some unseen accuracy intact).
In all cases, the harmonic mean and AUSUC metrics confirm that no calibration can rescue the model once CBEA is applied with sufficient strength.
% The trade-off is that the weakest CBEA-1 offers no additional advantage over clsA-1, but even moderate strength (CBEA-5 and -10) yields extreme failure.

\subsection{Results of NCPconA}
\label{sec:exp_NCPconA}

Table~\ref{table:exp_NCPconA} shows the effect of the NCPconA on the model’s concept prediction error. The key metric is the MSE of the predicted concepts for unseen ($MSE_u$) and seen ($MSE_s$) classes.
In the clean condition, for PSVMA, the unseen class $MSE_u$ is $2.38$ for AWA2, $0.57$ for CUB, and $1.74$ for SUN.
At NCPconA-1, these errors are identical to clean, reflecting the minimal perturbation.
At NCPconA-5, the errors grow substantially: AWA2 $MSE_u=3.59$ (v.s. $2.38$ clean), CUB $MSE_u=0.95$ (v.s. $0.57$), SUN $MSE_u=2.92$ (v.s. $1.74$).
By NCPconA-10, the attacks further increase error (AWA2 $MSE_u=3.91$, CUB $=1.00$, SUN $=3.25$).

Notably, seen-class errors ($MSE_s$) increase almost identically to unseen-class errors (e.g., PSVMA's AWA2 $MSE_s=3.81$ at NCPconA-10 v.s. $MSE_u=3.91$), indicating that NCPconA perturbs concept predictions across both classes equally.
Comparing datasets, CUB has much lower MSE in all cases (PSVMA's baseline $MSE_u=0.57$ rising to $1.00$ at NCPconA-10) than AWA2 or SUN (e.g., PSVMA's baseline is about $MSE_u=1.74$ rising to about $3.25$ on SUN).
This suggests that CUB’s attribute model is inherently more accurate, so the same attack strength yields smaller numerical MSE.
However, the relative degradation is similar: all three see roughly a $3-4\times$ increase in error at the strongest attack.
AWA2 and SUN, with larger attribute spaces and higher baseline error, show larger absolute increases.

NCPconA’s effectiveness lies in degrading the semantic consistency of the model.
However, it might change the class prediction.
Thus, we report the more class-presever concept attack NPconA in the next part.

\subsection{Results of CPconA}
\label{sec:exp_CPconA}

Table~\ref{table:exp_CPconA} reports the Class-Preserving Concept Attack (CPconA) results, which also target concept predictions but constrain the attack to preserve each input’s original class. Under CPconA-1 the concept errors remain at the clean levels ($MSE_u$ and $MSE_s$ unchanged). At CPconA-5, concept MSE increases modestly: for ReZSL on AWA2, unseen-class MSE rises to $2.21$ (from $1.60$ clean); CUB rises to $0.54$ (from $0.36$); SUN to $1.45$ (from $1.09$).
Further increasing the attack strength to CPconA-10 has virtually no additional effect (e.g., ReZSL on AWA2 $MSE_u=2.20$ at CPconA-10 vs $2.21$ at CPconA-5). This saturation indicates that CPconA’s constraint limits how much the concept predictions can deviate while still preserving the class label.
Cross-dataset, the patterns mirror NCPconA but on a smaller scale: CUB’s errors remain lowest (ReZSL's $MSE_u=0.54$ at CPconA-5), AWA2 intermediate, SUN slightly higher for seen classes ($MSE_s=1.32$ vs $1.01$ clean).
The modest magnitude of error increase ($\leq0.8$ additional MSE for AWA2) shows that CPconA is much less disruptive than NCPconA. The primary trend is that CPconA pushes concept predictions off their optimum, but only by a limited amount.
 
Importantly, CPconA is designed not to alter the final class predictions (non-class-preserving), so the ZSL classification accuracy would be unchanged. In terms of effectiveness and limitations, CPconA shows that constraining the attack to preserve class severely limits its potency. The model remains largely accurate at the concept level (final $MSE_s=2.20$ under CPconA-10 for AWA2 vs $4.41$ under NCPconA-10). This reflects a fundamental trade-off: to keep the class label fixed, the adversarial perturbation cannot arbitrarily distort the attribute space.
Thus, CPconA is effective only at introducing mild semantic noise, and this noise quickly saturates. In practice, CPconA demonstrates that ZSL models can be made robust to perturbations that preserve class, at the expense of attribute accuracy; but it is inherently weaker than the unconstrained concept attack.

\subsection{Visualization}
\label{sec:exp_vis}

\begin{figure*}[htbp]
    \centering
    \includegraphics[width=0.95\linewidth]{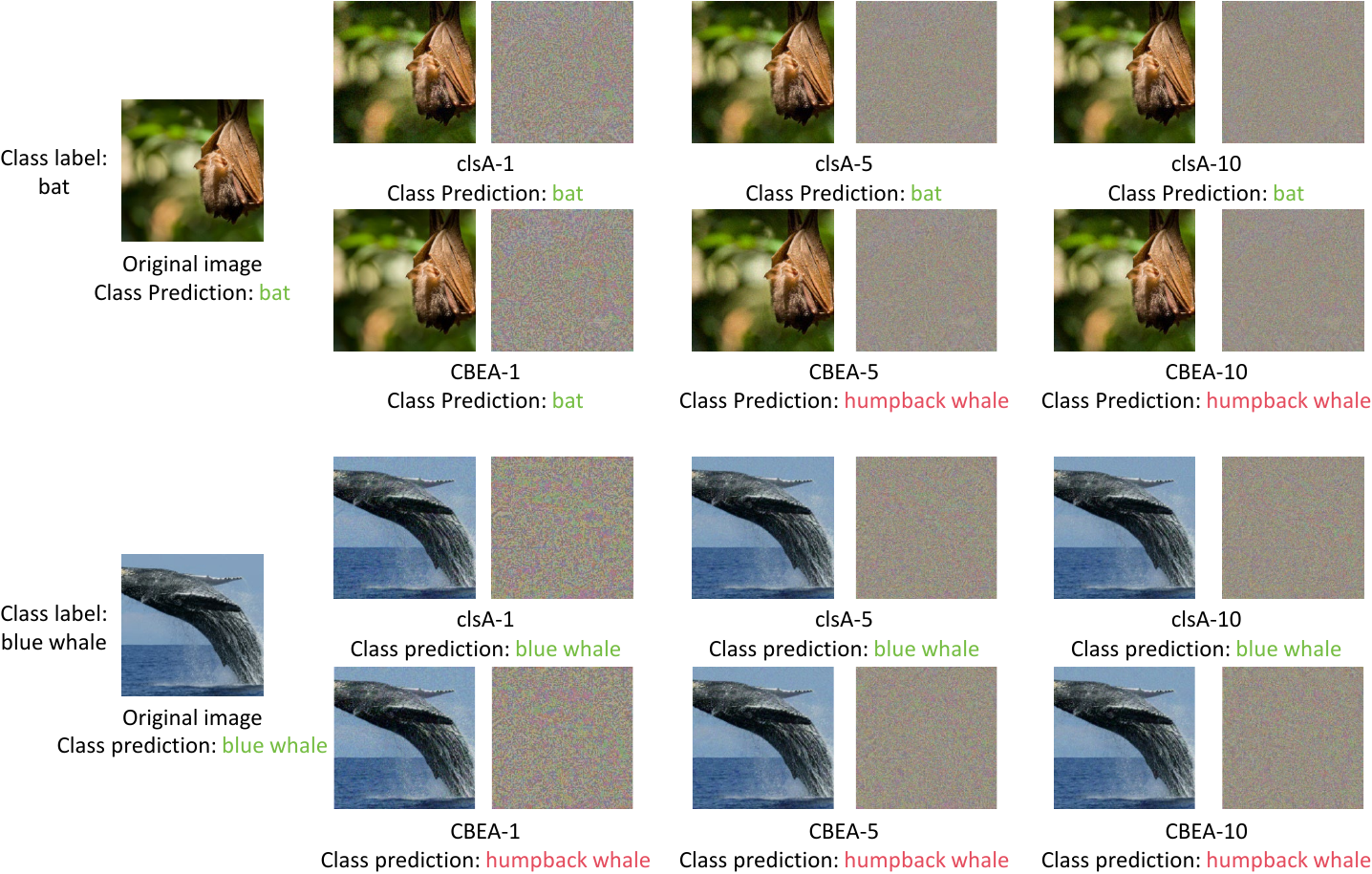}
    \caption{Two groups of adversarial examples and corresponding noises produced by our clsA and CBEA. The produced malicious noises are imperceptible. Moreover, we can find that our CBEA can attack successfully with fewer noise update. }
    \label{fig:adv_example1}
\end{figure*}

\begin{figure*}[htbp]
    \centering
    \includegraphics[width=0.95\linewidth]{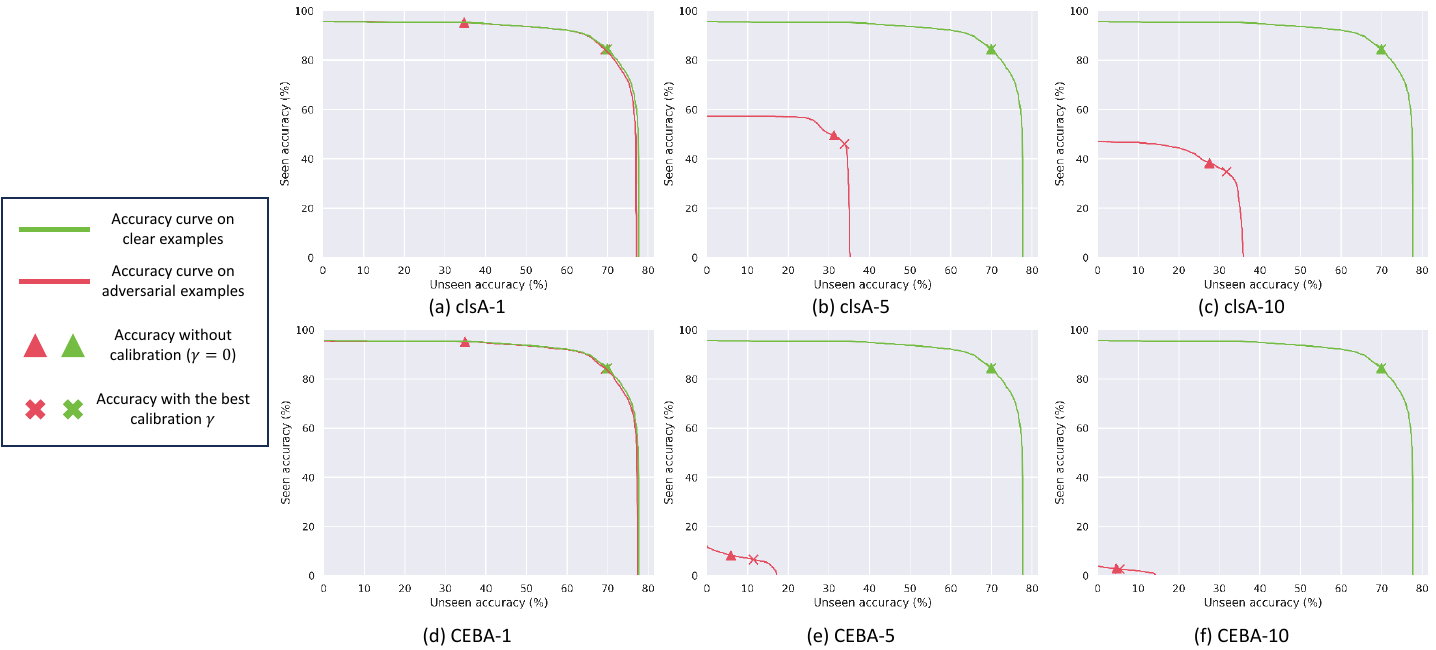}
    \caption{The AUSUC comparison about our plain clsA and CEBA  on PSVMA model. We can find that the adversarial examples produced from the plain clsA still remain a large accuracy area, while \textbf{our CEBA effectively removes almost accuracy area with enough noise update}. }
    \label{fig:AUSUC}
\end{figure*}

\begin{figure*}[htbp]
    \centering
    \includegraphics[width=0.95\linewidth]{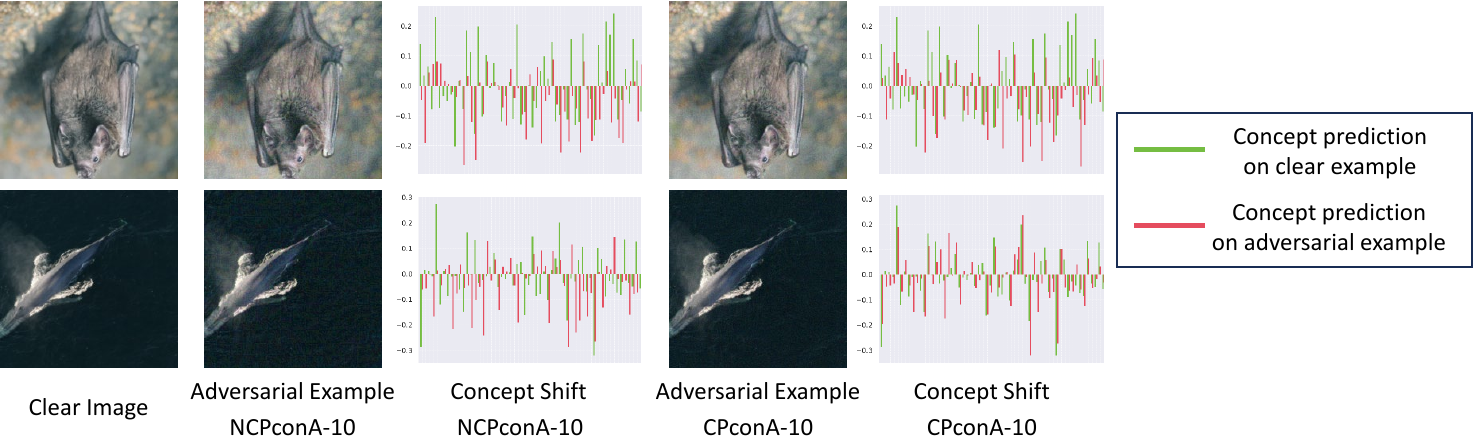}
    \caption{The example about our NCPconA-10 and CPconA-10. Minor noise on the images can also cause drastic changes in concept prediction, regardless of whether class prediction is affected or not. }
    \label{fig:adv_concept}
\end{figure*}
\textbf{Adversarial Example Visualization}
To further verify the effectiveness of our proposed class attack methods, we visualize the produced adversarial examples and corresponding noises of clsA and CEBA in Fig.~\ref{fig:adv_example1}. In addition to our adversarial examples looking almost similar to the original images, we notice two more observations: (1) less noise update steps produce more uneven noise distributions; For example, in clsA-1 and CBEA-1, the noise is more colorful than ones produces by clsA-10 and CBEA-10. That indicates the noise becomes more smoothing with more update steps. (2) Our CEBA could effectively change the classification prediction with smaller update steps than the plain clsA.

We also provide the AUSUC visualization in Fig.~\ref{fig:AUSUC}. We can find that clsA remains a relatively large area of AUSUC (The area surrounded by the red line). That means models can easily restore accurate class predictions by changing $\gamma$. In contrast, CEBA dramatically reduces the recoverable area. That means models are destroyed completely by CEBA.

\textbf{Perturbed Concept Visualization}
We also provide the visualization of adversarial examples and perturbed concepts of concept attacks in Fig.~\ref{fig:adv_concept}. As the concept attacks essentially are maximizing the difference between clear and noised concept predictions, we visualize the MSE per concept of CPconA and NCPconA. We can find that minor noise on the images can also cause drastic changes in concept prediction, regardless of whether class prediction is affected or not.
This may pose serious potential risks, especially in some fields involving important concept predictions, such as medical images and autonomous driving.

\section{Conclusion}
In this paper, we conducted the first systematic analysis on adversarial attacks/examples against ZSL models for both class and concept prediction with four attack methods (two for classes, two for concepts).
Specifically, we successfully disrupted the class prediction by the well-known non-target class attack (clsA) to analyze that current ZSL models are vulnerable to adversarial perturbations in ZSL settings.
However, we observe a spurious attack success of the well-known clsA in GZSL setting as a widely-used class calibration technique.
To address this, we propose the CBEA that directly enhances the gap between seen and unseen class probabilities.
Next, we introduced two concept-based attacks: CPconA and NCPconA that further reveal the fragility of ZSL models.
These show that concept predictions could also be easily changed, regardless of whether class prediction is affected or not.
Our findings highlight a significant performance gap between existing approaches, emphasizing the need for improved adversarial robustness in current ZSL models.

% \section*{Acknowledgments}
% This work was partially supported by National Natural Science Foundation of China under No. XXXX.
	
%{\appendices
%\section*{Proof of the First Zonklar Equation}
%Appendix one text goes here.
% You can choose not to have a title for an appendix if you want by leaving the argument blank
%\section*{Proof of the Second Zonklar Equation}
%Appendix two text goes here.}
	
% \newpage

%\section{References Section}
\bibliography{attackZSL}

@String(ICME = {Int. Conf. Multimedia and Expo})

@String(AAAI = {AAAI})

@String(ICME  =	{ICME})

@inproceedings{xie2019attentive,
	title={Attentive region embedding network for zero-shot learning},
	author={Xie, Guo-Sen and Liu, Li and Jin, Xiaobo and Zhu, Fan and Zhang, Zheng and Qin, Jie and Yao, Yazhou and Shao, Ling},
	booktitle={Proceedings of the IEEE/CVF Conference on Computer Vision and Pattern Recognition},
	pages={9384--9393},
	year={2019}
}

@inproceedings{xie2020region,
	title={Region graph embedding network for zero-shot learning},
	author={Xie, Guo-Sen and Liu, Li and Zhu, Fan and Zhao, Fang and Zhang, Zheng and Yao, Yazhou and Qin, Jie and Shao, Ling},
	booktitle={European Conference on Computer Vision},
	pages={562--580},
	year={2020},
	organization={Springer}
}

@inproceedings{xu2020attribute,
	title={Attribute Prototype Network for Zero-Shot Learning},
	author={Xu, Wenjia and Xian, Yongqin and Wang, Jiuniu and Schiele, Bernt and Akata, Zeynep},
	booktitle={NeurIPS},
	year={2020}
}

@inproceedings{liu2021goal,
	title={Goal-oriented gaze estimation for zero-shot learning},
	author={Liu, Yang and Zhou, Lei and Bai, Xiao and Huang, Yifei and Gu, Lin and Zhou, Jun and Harada, Tatsuya},
	booktitle={Proceedings of the IEEE/CVF Conference on Computer Vision and Pattern Recognition},
	pages={3794--3803},
	year={2021}
}

@inproceedings{chen2022transzero,
	title={TransZero: Attribute-guided Transformer for Zero-Shot Learning},
	author={Chen, Shiming and Hong, Ziming and Liu, Yang and Xie, Guo-sen and Sun, Baigui and Li, Hao and Peng, Qinmu and Lu, Ke and You, Xinge},
	booktitle={AAAI},
	year={2022}
}

@inproceedings{xian2018feature,
	title={Feature generating networks for zero-shot learning},
	author={Xian, Yongqin and Lorenz, Tobias and Schiele, Bernt and Akata, Zeynep},
	booktitle={Proceedings of the IEEE conference on computer vision and pattern recognition},
	pages={5542--5551},
	year={2018}
}

@article{ye2021disentangling,
	title={Disentangling Semantic-to-visual Confusion for Zero-shot Learning},
	author={Ye, Zihan and Hu, Fuyuan and Lyu, Fan and Li, Linyan and Huang, Kaizhu},
	journal={IEEE Transactions on Multimedia},
	year={2021},
	publisher={IEEE}
}

@article{chen2021hsva,
	title={Hsva: Hierarchical semantic-visual adaptation for zero-shot learning},
	author={Chen, Shiming and Xie, Guosen and Liu, Yang and Peng, Qinmu and Sun, Baigui and Li, Hao and You, Xinge and Shao, Ling},
	journal={Advances in Neural Information Processing Systems},
	volume={34},
	year={2021}
}

@inproceedings{chen2021free,
	title={Free: Feature refinement for generalized zero-shot learning},
	author={Chen, Shiming and Wang, Wenjie and Xia, Beihao and Peng, Qinmu and You, Xinge and Zheng, Feng and Shao, Ling},
	booktitle={Proceedings of the IEEE/CVF international conference on computer vision},
	pages={122--131},
	year={2021}
}

@inproceedings{min2020domain,
	title={Domain-aware visual bias eliminating for generalized zero-shot learning},
	author={Min, Shaobo and Yao, Hantao and Xie, Hongtao and Wang, Chaoqun and Zha, Zheng-Jun and Zhang, Yongdong},
	booktitle={Proceedings of the IEEE/CVF Conference on Computer Vision and Pattern Recognition},
	pages={12664--12673},
	year={2020}
}

@article{xian2018zero,
	title={Zero-shot learning—a comprehensive evaluation of the good, the bad and the ugly},
	author={Xian, Yongqin and Lampert, Christoph H and Schiele, Bernt and Akata, Zeynep},
	journal={IEEE transactions on pattern analysis and machine intelligence},
	volume={41},
	number={9},
	pages={2251--2265},
	year={2018},
	publisher={IEEE}
}

@inproceedings{chao2016empirical,
	title={An empirical study and analysis of generalized zero-shot learning for object recognition in the wild},
	author={Chao, Wei-Lun and Changpinyo, Soravit and Gong, Boqing and Sha, Fei},
	booktitle={European conference on computer vision},
	pages={52--68},
	year={2016},
	organization={Springer}
}

@article{wah2011caltech,
	title={The caltech-ucsd birds-200-2011 dataset},
	author={Wah, Catherine and Branson, Steve and Welinder, Peter and Perona, Pietro and Belongie, Serge},
	year={2011},
	publisher={California Institute of Technology}
}

@inproceedings{ye2019sr,
	title={Sr-gan: Semantic rectifying generative adversarial network for zero-shot learning},
	author={Ye, Zihan and Lyu, Fan and Li, Linyan and Fu, Qiming and Ren, Jinchang and Hu, Fuyuan},
	booktitle={2019 IEEE International Conference on Multimedia and Expo (ICME)},
	pages={85--90},
	year={2019},
	organization={IEEE}
}

@inproceedings{shen2020invertible,
	title={Invertible zero-shot recognition flows},
	author={Shen, Yuming and Qin, Jie and Huang, Lei and Liu, Li and Zhu, Fan and Shao, Ling},
	booktitle={European Conference on Computer Vision},
	pages={614--631},
	year={2020},
	organization={Springer}
}

@inproceedings{li2018discriminative,
	title={Discriminative learning of latent features for zero-shot recognition},
	author={Li, Yan and Zhang, Junge and Zhang, Jianguo and Huang, Kaiqi},
	booktitle={Proceedings of the IEEE Conference on Computer Vision and Pattern Recognition},
	pages={7463--7471},
	year={2018}
}

@article{lampert2013attribute,
	title={Attribute-based classification for zero-shot visual object categorization},
	author={Lampert, Christoph H and Nickisch, Hannes and Harmling, Stefan},
	journal={IEEE transactions on pattern analysis and machine intelligence},
	volume={36},
	number={3},
	pages={453--465},
	year={2013},
	publisher={IEEE}
}

@inproceedings{patterson2012sun,
	title={Sun attribute database: Discovering, annotating, and recognizing scene attributes},
	author={Patterson, Genevieve and Hays, James},
	booktitle={2012 IEEE Conference on Computer Vision and Pattern Recognition},
	pages={2751--2758},
	year={2012},
	organization={IEEE}
}

@article{du2022boosting,
  title={Boosting Zero-shot Learning via Contrastive Optimization of Attribute Representations},
  author={Du, Yu and Shi, Miaojing and Wei, Fangyun and Li, Guoqi},
  journal={arXiv preprint arXiv:2207.03824},
  year={2022}
}

@inproceedings{pennington2014glove,
  title={Glove: Global vectors for word representation},
  author={Pennington, Jeffrey and Socher, Richard and Manning, Christopher D},
  booktitle={Proceedings of the 2014 conference on empirical methods in natural language processing (EMNLP)},
  pages={1532--1543},
  year={2014}
}

@inproceedings{reed2016learning,
	title={Learning deep representations of fine-grained visual descriptions},
	author={Reed, Scott and Akata, Zeynep and Lee, Honglak and Schiele, Bernt},
	booktitle={Proceedings of the IEEE conference on computer vision and pattern recognition},
	pages={49--58},
	year={2016}
}

@article{ye2023rebalanced,
  title={Rebalanced zero-shot learning},
  author={Ye, Zihan and Yang, Guanyu and Jin, Xiaobo and Liu, Youfa and Huang, Kaizhu},
  journal={IEEE Transactions on Image Processing},
  year={2023},
  publisher={IEEE}
}

@article{chen2022transzero++,
  title={TransZero++: Cross attribute-guided transformer for zero-shot learning},
  author={Chen, Shiming and Hong, Ziming and Hou, Wenjin and Xie, Guo-Sen and Song, Yibing and Zhao, Jian and You, Xinge and Yan, Shuicheng and Shao, Ling},
  journal={IEEE transactions on pattern analysis and machine intelligence},
  year={2022},
  publisher={IEEE}
}

@article{naeem2022i2dformer,
  title={I2dformer: Learning image to document attention for zero-shot image classification},
  author={Naeem, Muhammad Ferjad and Xian, Yongqin and Gool, Luc V and Tombari, Federico},
  journal={Advances in Neural Information Processing Systems},
  volume={35},
  pages={12283--12294},
  year={2022}
}

@inproceedings{liu2023progressive,
  title={Progressive semantic-visual mutual adaption for generalized zero-shot learning},
  author={Liu, Man and Li, Feng and Zhang, Chunjie and Wei, Yunchao and Bai, Huihui and Zhao, Yao},
  booktitle={Proceedings of the IEEE/CVF Conference on Computer Vision and Pattern Recognition},
  pages={15337--15346},
  year={2023}
}

@inproceedings{chen2024progressive,
  title={Progressive Semantic-Guided Vision Transformer for Zero-Shot Learning},
  author={Chen, Shiming and Hou, Wenjin and Khan, Salman and Khan, Fahad Shahbaz},
  booktitle={Proceedings of the IEEE/CVF Conference on Computer Vision and Pattern Recognition},
  pages={23964--23974},
  year={2024}
}

@inproceedings{chen2024causal,
  title={Causal Visual-semantic Correlation for Zero-shot Learning},
  author={Chen, Shuhuang and Fu, Dingjie and Chen, Shiming and Hou, Wenjin and You, Xinge and others},
  booktitle={ACM Multimedia 2024},
  year={2024}
}

@inproceedings{naeem2023i2mvformer,
  title={I2mvformer: Large language model generated multi-view document supervision for zero-shot image classification},
  author={Naeem, Muhammad Ferjad and Khan, Muhammad Gul Zain Ali and Xian, Yongqin and Afzal, Muhammad Zeshan and Stricker, Didier and Van Gool, Luc and Tombari, Federico},
  booktitle={Proceedings of the IEEE/CVF Conference on Computer Vision and Pattern Recognition},
  pages={15169--15179},
  year={2023}
}

@inproceedings{xu2022vgse,
  title={Vgse: Visually-grounded semantic embeddings for zero-shot learning},
  author={Xu, Wenjia and Xian, Yongqin and Wang, Jiuniu and Schiele, Bernt and Akata, Zeynep},
  booktitle={Proceedings of the IEEE/CVF Conference on Computer Vision and Pattern Recognition},
  pages={9316--9325},
  year={2022}
}

@inproceedings{yang2023language,
  title={Language in a bottle: Language model guided concept bottlenecks for interpretable image classification},
  author={Yang, Yue and Panagopoulou, Artemis and Zhou, Shenghao and Jin, Daniel and Callison-Burch, Chris and Yatskar, Mark},
  booktitle={Proceedings of the IEEE/CVF Conference on Computer Vision and Pattern Recognition},
  pages={19187--19197},
  year={2023}
}

@inproceedings{hou2025zeromamba,
  title={ZeroMamba: Exploring visual state space model for zero-shot learning},
  author={Hou, Wenjin and Fu, Dingjie and Li, Kun and Chen, Shiming and Fan, Hehe and Yang, Yi},
  booktitle={Proceedings of the AAAI Conference on Artificial Intelligence},
  volume={39},
  number={4},
  pages={3527--3535},
  year={2025}
}

@inproceedings{
ridnik2021imagenetk,
title={ImageNet-21K Pretraining for the Masses},
author={Tal Ridnik and Emanuel Ben-Baruch and Asaf Noy and Lihi Zelnik-Manor},
booktitle={Thirty-fifth Conference on Neural Information Processing Systems Datasets and Benchmarks Track (Round 1)},
year={2021},
}

@article{krizhevsky2012imagenet,
  title={Imagenet classification with deep convolutional neural networks},
  author={Krizhevsky, Alex and Sutskever, Ilya and Hinton, Geoffrey E},
  journal={Advances in neural information processing systems},
  volume={25},
  year={2012}
}

@article{wang2020generalizing,
  title={Generalizing from a few examples: A survey on few-shot learning},
  author={Wang, Yaqing and Yao, Quanming and Kwok, James T and Ni, Lionel M},
  journal={ACM computing surveys (csur)},
  volume={53},
  number={3},
  pages={1--34},
  year={2020},
  publisher={ACM New York, NY, USA}
}

@inproceedings{lampert2009learning,
  title={Learning to detect unseen object classes by between-class attribute transfer},
  author={Lampert, Christoph H and Nickisch, Hannes and Harmeling, Stefan},
  booktitle={2009 IEEE conference on computer vision and pattern recognition},
  pages={951--958},
  year={2009},
  organization={IEEE}
}

@inproceedings{zhang2017learning,
  title={Learning a deep embedding model for zero-shot learning},
  author={Zhang, Li and Xiang, Tao and Gong, Shaogang},
  booktitle={Proceedings of the IEEE conference on computer vision and pattern recognition},
  pages={2021--2030},
  year={2017}
}

@inproceedings{guo2023graph,
  title={Graph knows unknowns: Reformulate zero-shot learning as sample-level graph recognition},
  author={Guo, Jingcai and Guo, Song and Zhou, Qihua and Liu, Ziming and Lu, Xiaocheng and Huo, Fushuo},
  booktitle={Proceedings of the AAAI conference on artificial intelligence},
  volume={37},
  number={6},
  pages={7775--7783},
  year={2023}
}

@article{zhang2019adversarial,
  title={Adversarial examples: Opportunities and challenges},
  author={Zhang, Jiliang and Li, Chen},
  journal={IEEE transactions on neural networks and learning systems},
  volume={31},
  number={7},
  pages={2578--2593},
  year={2019},
  publisher={IEEE}
}

@article{han2023interpreting,
  title={Interpreting adversarial examples in deep learning: A review},
  author={Han, Sicong and Lin, Chenhao and Shen, Chao and Wang, Qian and Guan, Xiaohong},
  journal={ACM Computing Surveys},
  volume={55},
  number={14s},
  pages={1--38},
  year={2023},
  publisher={ACM New York, NY}
}

@inproceedings{chen2023zero,
  title={Zero-shot learning by harnessing adversarial samples},
  author={Chen, Zhi and Zhang, Pengfei and Li, Jingjing and Wang, Sen and Huang, Zi},
  booktitle={Proceedings of the 31st ACM international conference on multimedia},
  pages={4138--4146},
  year={2023}
}

@article{yucel2022robust,
  title={How robust are discriminatively trained zero-shot learning models?},
  author={Yucel, Mehmet Kerim and Cinbis, Ramazan Gokberk and Duygulu, Pinar},
  journal={Image and Vision Computing},
  volume={119},
  pages={104392},
  year={2022},
  publisher={Elsevier}
}

@inproceedings{yucel2020deep,
  title={A deep dive into adversarial robustness in zero-shot learning},
  author={Yucel, Mehmet Kerim and Cinbis, Ramazan Gokberk and Duygulu, Pinar},
  booktitle={European Conference on Computer Vision},
  pages={3--21},
  year={2020},
  organization={Springer}
}

@article{zhang2023atzsl,
  title={ATZSL: Defensive Zero-Shot Recognition in the Presence of Adversaries},
  author={Zhang, Xingxing and Gui, Shupeng and Jin, Jian and Zhu, Zhenfeng and Zhao, Yao},
  journal={IEEE Transactions on Multimedia},
  volume={26},
  pages={15--27},
  year={2023},
  publisher={IEEE}
}

@inproceedings{mahapatra2021medical,
  title={Medical image classification using generalized zero shot learning},
  author={Mahapatra, Dwarikanath and Bozorgtabar, Behzad and Ge, Zongyuan},
  booktitle={Proceedings of the IEEE/CVF international conference on computer vision},
  pages={3344--3353},
  year={2021}
}

@article{rezaei2020zero,
  title={Zero-shot learning and its applications from autonomous vehicles to COVID-19 diagnosis: A review},
  author={Rezaei, Mahdi and Shahidi, Mahsa},
  journal={Intelligence-based medicine},
  volume={3},
  pages={100005},
  year={2020},
  publisher={Elsevier}
}

@article{mahapatra2022self,
  title={Self-supervised generalized zero shot learning for medical image classification using novel interpretable saliency maps},
  author={Mahapatra, Dwarikanath and Ge, Zongyuan and Reyes, Mauricio},
  journal={IEEE Transactions on Medical Imaging},
  volume={41},
  number={9},
  pages={2443--2456},
  year={2022},
  publisher={IEEE}
}

@article{gupta2023generative,
  title={Generative multi-label zero-shot learning},
  author={Gupta, Akshita and Narayan, Sanath and Khan, Salman and Khan, Fahad Shahbaz and Shao, Ling and Van De Weijer, Joost},
  journal={IEEE Transactions on Pattern Analysis and Machine Intelligence},
  volume={45},
  number={12},
  pages={14611--14624},
  year={2023},
  publisher={IEEE}
}

@inproceedings{madry2018towards,
title={Towards Deep Learning Models Resistant to Adversarial Attacks},
author={Aleksander Madry and Aleksandar Makelov and Ludwig Schmidt and Dimitris Tsipras and Adrian Vladu},
booktitle={International Conference on Learning Representations},
year={2018},
}

@inproceedings{zhang2025rp,
  title={RP-PGD: Boosting Segmentation Robustness with a Region-and-Prototype Based Adversarial Attack},
  author={Zhang, Yuxuan and Shi, Zhenbo and Wang, Shuchang and Yang, Wei and Wang, Shaowei and Xue, Yinxing},
  booktitle={Proceedings of the AAAI Conference on Artificial Intelligence},
  volume={39},
  number={10},
  pages={10338--10347},
  year={2025}
}

@article{chen2023explanatory,
  title={Explanatory object part aggregation for zero-shot learning},
  author={Chen, Xin and Deng, Xiaoling and Lan, Yubin and Long, Yongbing and Weng, Jian and Liu, Zhiquan and Tian, Qi},
  journal={IEEE Transactions on Pattern Analysis and Machine Intelligence},
  volume={46},
  number={2},
  pages={851--868},
  year={2023},
  publisher={IEEE}
}

@inproceedings{ye2025zerodiff,
title={ZeroDiff: Solidified Visual-semantic Correlation in Zero-Shot Learning},
author={Zihan Ye and Shreyank N Gowda and Shiming Chen and Xiaowei Huang and Haotian Xu and Fahad Shahbaz Khan and Yaochu Jin and Kaizhu Huang and Xiaobo Jin},
booktitle={The Thirteenth International Conference on Learning Representations},
year={2025},
}

@article{chen2022gsmflow,
  title={GSMFlow: Generation shifts mitigating flow for generalized zero-shot learning},
  author={Chen, Zhi and Luo, Yadan and Wang, Sen and Li, Jingjing and Huang, Zi},
  journal={IEEE Transactions on Multimedia},
  volume={25},
  pages={5374--5385},
  year={2022},
  publisher={IEEE}
}

@article{gao2025self,
  title={Self-Assembled Generative Framework for Generalized Zero-Shot Learning},
  author={Gao, Mengyu and Dong, Qiulei},
  journal={IEEE Transactions on Image Processing},
  year={2025},
  publisher={IEEE}
}

@inproceedings{akata2016multi,
  title={Multi-cue zero-shot learning with strong supervision},
  author={Akata, Zeynep and Malinowski, Mateusz and Fritz, Mario and Schiele, Bernt},
  booktitle={Proceedings of the IEEE conference on computer vision and pattern recognition},
  pages={59--68},
  year={2016}
}

@article{du2023boosting,
  title={Boosting zero-shot learning via contrastive optimization of attribute representations},
  author={Du, Yu and Shi, Miaojing and Wei, Fangyun and Li, Guoqi},
  journal={IEEE Transactions on Neural Networks and Learning Systems},
  year={2023},
  publisher={IEEE}
}

@article{goodfellow2014explaining,
  title={Explaining and harnessing adversarial examples},
  author={Goodfellow, Ian J and Shlens, Jonathon and Szegedy, Christian},
  journal={arXiv preprint arXiv:1412.6572},
  year={2014}
}

@inproceedings{kurakin2017adversarial,
title={Adversarial Machine Learning at Scale},
author={Alexey Kurakin and Ian J. Goodfellow and Samy Bengio},
booktitle={International Conference on Learning Representations},
year={2017},
}

@article{su2019one,
  title={One pixel attack for fooling deep neural networks},
  author={Su, Jiawei and Vargas, Danilo Vasconcellos and Sakurai, Kouichi},
  journal={IEEE Transactions on Evolutionary Computation},
  volume={23},
  number={5},
  pages={828--841},
  year={2019},
  publisher={IEEE}
}

@inproceedings{eykholt2018robust,
  title={Robust physical-world attacks on deep learning visual classification},
  author={Eykholt, Kevin and Evtimov, Ivan and Fernandes, Earlence and Li, Bo and Rahmati, Amir and Xiao, Chaowei and Prakash, Atul and Kohno, Tadayoshi and Song, Dawn},
  booktitle={Proceedings of the IEEE conference on computer vision and pattern recognition},
  pages={1625--1634},
  year={2018}
}

@inproceedings{le2024towards,
  title={Towards Robust Saliency Maps},
  author={Le, Nham and Gurfinkel, Arie and Si, Xujie and Geng, Chuqin},
  booktitle={The 16th Asian Conference on Machine Learning (Conference Track)},
  year={2024}
}

@inproceedings{lakkaraju2020robust,
  title={Robust and stable black box explanations},
  author={Lakkaraju, Himabindu and Arsov, Nino and Bastani, Osbert},
  booktitle={International conference on machine learning},
  pages={5628--5638},
  year={2020},
  organization={PMLR}
}

@inproceedings{huai2022towards,
  title={Towards automating model explanations with certified robustness guarantees},
  author={Huai, Mengdi and Liu, Jinduo and Miao, Chenglin and Yao, Liuyi and Zhang, Aidong},
  booktitle={Proceedings of the AAAI Conference on Artificial Intelligence},
  volume={36},
  number={6},
  pages={6935--6943},
  year={2022}
}

@inproceedings{kim2005kernel,
  title={Kernel based intrusion detection system},
  author={Kim, Byung-joo and Kim, Il-kon},
  booktitle={Fourth Annual ACIS International Conference on Computer and Information Science (ICIS'05)},
  pages={13--18},
  year={2005},
  organization={IEEE}
}

@inproceedings{marzi2018sparsity,
  title={Sparsity-based defense against adversarial attacks on linear classifiers},
  author={Marzi, Zhinus and Gopalakrishnan, Soorya and Madhow, Upamanyu and Pedarsani, Ramtin},
  booktitle={2018 IEEE International Symposium on Information Theory (ISIT)},
  pages={31--35},
  year={2018},
  organization={IEEE}
}

@article{akhtar2018threat,
  title={Threat of adversarial attacks on deep learning in computer vision: A survey},
  author={Akhtar, Naveed and Mian, Ajmal},
  journal={Ieee Access},
  volume={6},
  pages={14410--14430},
  year={2018},
  publisher={IEEE}
}

@article{szegedy2013intriguing,
  title={Intriguing properties of neural networks},
  author={Szegedy, Christian and Zaremba, Wojciech and Sutskever, Ilya and Bruna, Joan and Erhan, Dumitru and Goodfellow, Ian and Fergus, Rob},
  journal={arXiv preprint arXiv:1312.6199},
  year={2013}
}

@article{papernot2016transferability,
  title={Transferability in machine learning: from phenomena to black-box attacks using adversarial samples},
  author={Papernot, Nicolas and McDaniel, Patrick and Goodfellow, Ian},
  journal={arXiv preprint arXiv:1605.07277},
  year={2016}
}

@article{li2024spiking,
  title={Spiking tucker fusion transformer for audio-visual zero-shot learning},
  author={Li, Wenrui and Wang, Penghong and Xiong, Ruiqin and Fan, Xiaopeng},
  journal={IEEE Transactions on Image Processing},
  year={2024},
  publisher={IEEE}
}

@article{wu2024prototype,
  title={Prototype-augmented self-supervised generative network for generalized zero-shot learning},
  author={Wu, Jiamin and Zhang, Tianzhu and Zha, Zheng-Jun and Luo, Jiebo and Zhang, Yongdong and Wu, Feng},
  journal={IEEE Transactions on Image Processing},
  volume={33},
  pages={1938--1951},
  year={2024},
  publisher={IEEE}
}

@article{lu2024pamk,
  title={Pamk: Prototype augmented multi-teacher knowledge transfer network for continual zero-shot learning},
  author={Lu, Junxin and Sun, Shiliang},
  journal={IEEE Transactions on Image Processing},
  volume={33},
  pages={3353--3368},
  year={2024},
  publisher={IEEE}
}

@inproceedings{cui2024robustness,
  title={On the robustness of large multimodal models against image adversarial attacks},
  author={Cui, Xuanming and Aparcedo, Alejandro and Jang, Young Kyun and Lim, Ser-Nam},
  booktitle={Proceedings of the IEEE/CVF Conference on Computer Vision and Pattern Recognition},
  pages={24625--24634},
  year={2024}
}

@article{jiang2025survey,
  title={Survey of adversarial robustness in multimodal large language models},
  author={Jiang, Chengze and Wang, Zhuangzhuang and Dong, Minjing and Gui, Jie},
  journal={arXiv preprint arXiv:2503.13962},
  year={2025}
}

@article{zhou2024revisiting,
  title={Revisiting the adversarial robustness of vision language models: a multimodal perspective},
  author={Zhou, Wanqi and Bai, Shuanghao and Mandic, Danilo P and Zhao, Qibin and Chen, Badong},
  journal={arXiv preprint arXiv:2404.19287},
  year={2024}
}

@article{xu2023multimodal,
  title={Multimodal learning with transformers: A survey},
  author={Xu, Peng and Zhu, Xiatian and Clifton, David A},
  journal={IEEE Transactions on Pattern Analysis and Machine Intelligence},
  volume={45},
  number={10},
  pages={12113--12132},
  year={2023},
  publisher={IEEE}
}

@article{yuan2025survey,
  title={A survey of multimodal learning: Methods, applications, and future},
  author={Yuan, Yuan and Li, Zhaojian and Zhao, Bin},
  journal={ACM Computing Surveys},
  volume={57},
  number={7},
  pages={1--34},
  year={2025},
  publisher={ACM New York, NY}
}
\bibliographystyle{IEEEtran}

\vspace{11pt}

\vfill
\newpage

\end{document}